\documentclass[10pt,twocolumn,letterpaper]{article}
\usepackage[T1]{fontenc}
\usepackage[pagenumbers]{cvpr}
\input{preamble-arxiv}
\definecolor{cvprblue}{rgb}{0.21,0.49,0.74}
\usepackage[breaklinks,colorlinks,allcolors=cvprblue]{hyperref}
\hypersetup{pdftitle={TaF-Foundation: Wrench-Grounded Cross-Sensor Tactile Representations}}
\title{TaF-Foundation: Wrench-Grounded Cross-Sensor Tactile Representations}
\author{
    Jiaming Jiang\textsuperscript{1,3$\star$},
    Yuzhe Huang\textsuperscript{2$\star$},
    Baokun Yang\textsuperscript{3,4},
    Hao Liang\textsuperscript{3},
    Li Gao\textsuperscript{5}, \\
    Haoran Guo\textsuperscript{1},
    Pei Lin\textsuperscript{1},
    Shengcheng Luo\textsuperscript{1},
    Chenxi Xiao\textsuperscript{1$\dagger$},
    Wanlin Li\textsuperscript{3$\dagger$},
    Ziyuan Jiao\textsuperscript{2$\dagger$} \\[0.5em]
    \textsuperscript{1}ShanghaiTech University \quad
    \textsuperscript{2}Beihang University \quad \\
    \textsuperscript{3}Beijing Institute for General Artificial Intelligence \quad 
    \textsuperscript{4}Beijing Jiaotong University \quad
    \textsuperscript{5}AMAP, Alibaba \quad \\[0.5em]
    \textsuperscript{$\star$}Equal contributors \quad
    \textsuperscript{$\dagger$}Corresponding authors 
}
\date{}
\begin{document}
\maketitle
\begin{abstract}
Tactile sensing is essential for contact-rich manipulation, yet heterogeneous sensing mechanisms and observation formats hinder cross-hardware transfer. Since physical contact is governed by Newtonian mechanics, we ground tactile representations in six-dimensional wrenches, establishing a common physical reference across sensing mechanisms. Specifically, we pair tactile observations from all sensors with wrench measurements, enabling shared physical supervision without requiring direct tactile-to-tactile pairing between sensors. To support learning at scale, we develop a dedicated acquisition platform and collect over 100 million paired tactile observations and wrench measurements from 24 sensors spanning four families of sensing principles. Building on this dataset, our framework combines sensor-specific input adaptation with a shared tactile encoder. By learning to reconstruct wrenches, the encoder maps observations from different sensors into a shared, continuous feature space grounded in contact physics. To leverage these representations for manipulation, we introduce TaF-Policy, which integrates tactile features with visual observations and robot states to generate action chunks. We evaluate the learned representations through wrench estimation, cross-sensor retrieval, few-shot adaptation to unseen sensors, and robotic manipulation. On the core benchmark, our method reduces single-frame force RMSE by 17.1 \% and improves cross-sensor retrieval R@1 by 13.11 percentage points over the strongest baseline. In manipulation experiments, TaF-Policy achieves average success rates of 51.33\% across three simulated tasks and 46.25\% across four real-world tasks, compared with 47.33\% and 23.75\%, respectively, for the strongest baselines evaluated on the same task sets. Project page: \href{https://mrjiangjm.github.io/TaF-Foundation/}{\textcolor{blue}{https://mrjiangjm.github.io/TaF-Foundation/}}.
\end{abstract}

\section{Introduction}
\label{sec:intro}

\begin{figure*}[t!]
    \centering
    \includegraphics[width=\linewidth]{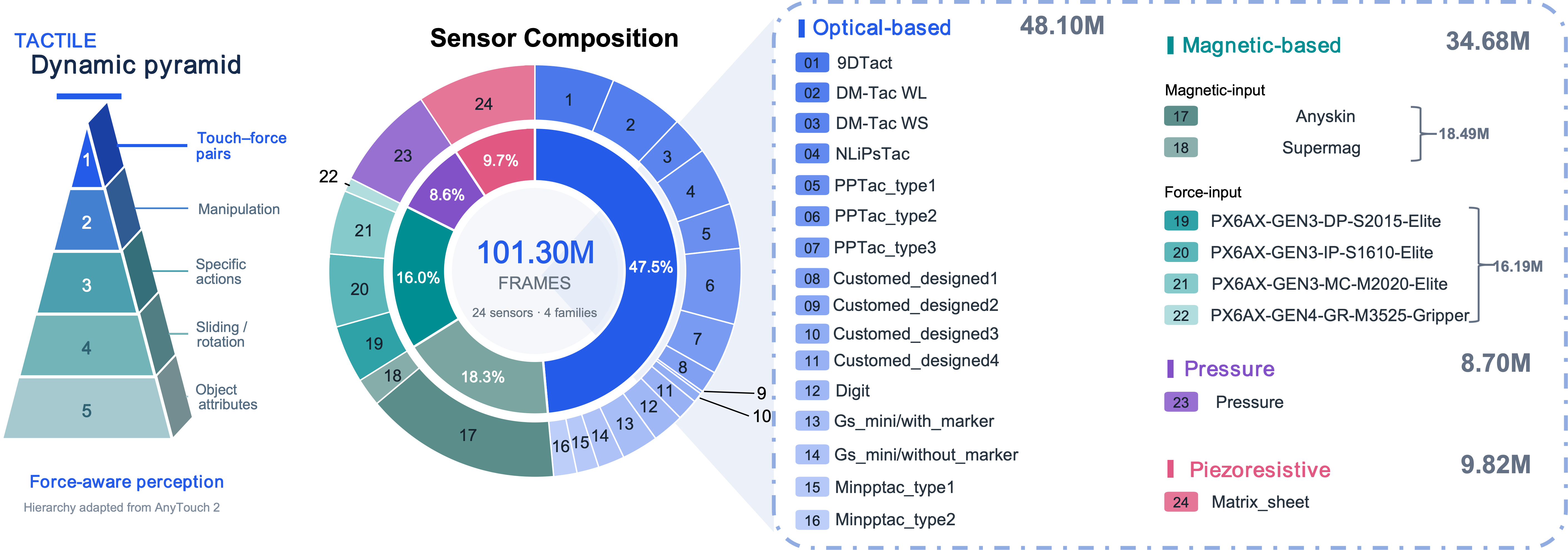}
    \caption{
        Dataset coverage across heterogeneous tactile sensors.
        Left: a hierarchy of tactile learning tasks, adapted from
        AnyTouch~2~\cite{feng2026anytouch}.
        Center: the distribution of paired observations across
        24 sensors in four families; the inner ring shows family
        proportions, and the outer ring shows individual sensors.
        Right: sensor inventory and observation counts by family.
    }

    \label{fig:dataset_coverage}
\end{figure*}

Reliable manipulation requires robots to understand both what they see and how they physically interact with the world~\cite{billard2019trends}. Vision provides rich information about objects and scenes, but contact states are often difficult to infer from visual appearance alone~\cite{mao2024multimodal,li2020review}. Tactile sensing therefore complements vision by capturing physical interactions, making it particularly valuable for contact-rich manipulation~\cite{lin2026dexmove,zhao2025embedding,zhao2024tac}. However, tactile sensors rely on heterogeneous sensing mechanisms and produce contact responses in different formats~\cite{yuan2017gelsight,bhirangi2025anyskin,huang2026flexitac,lin2025pp,qin2026nlipscalib,hou2025supermag}. This heterogeneity makes it difficult to integrate and reuse tactile data across sensors and tasks, limiting the scalability of tactile representation learning.

To improve the scalability and reusability of tactile representations, recent work has explored several representation learning strategies. Self-supervised learning with shared architectures provides a common framework for learning reusable features across tactile observations~\cite{higuera2024sparsh,zhao2024transferable}. To explicitly address sensor heterogeneity, cross-sensor alignment methods learn shared features from paired tactile observations collected under matched contact conditions~\cite{bi2026heterogeneous,park2026tactx}. More recently, several pioneering works have explored grounding tactile representations in physical contact properties, particularly force-related information~\cite{chen2026uniforce,feng2026anytouch,huang2026taf}. Despite this progress, existing approaches remain limited in sensor diversity and scale, leaving generalization to heterogeneous and unseen sensor hardware insufficiently explored. Moreover, independently collected tactile datasets lack a shared physical reference for comparing contact states across sensors, making it difficult to establish a physically meaningful metric space for representation learning. Therefore, we study an unanswered question: can large-scale supervision from directly measured physical quantities enable tactile representations to generalize across heterogeneous sensors, including unseen hardware?

Our key insight is that although different sensing mechanisms capture distinct manifestations of contact, the underlying physical interactions obey the same Newtonian mechanics. We therefore propose grounding tactile representations in six-dimensional wrenches, which quantify net contact forces and torques and characterize the mechanical interactions underlying manipulation. By independently pairing observations from each sensor with directly measured wrenches, we establish a shared physical supervision signal without requiring coordinated data collection across sensor pairs. This provides a common quantitative reference for comparing interactions across heterogeneous sensing modalities. 

To support large-scale learning, we develop a hardware platform that synchronously records tactile observations and six-dimensional wrenches, collecting over 100 million tactile--wrench pairs from 24 sensors across four sensing families: optical, magnetic, pressure-based, and piezoresistive (Fig.~\ref{fig:dataset_coverage}). We also establish a benchmark for evaluating representation quality and cross-sensor transferability. Building on this dataset, we develop a unified tactile encoder that maps heterogeneous observations into a shared continuous feature space. Sensor-specific input adapters accommodate different observation formats, while a shared Transformer aggregates tokens through a learnable physical query. We align tactile embeddings with targets from a momentum-updated wrench encoder and use a shared decoder to reconstruct measured wrenches, grounding cross-sensor representations in physical interactions. For downstream manipulation, we introduce TaF-Policy, which combines tactile features, visual observations, and robot states to generate action chunks.

Comprehensive benchmarks were conducted using learned representation. Our encoder reduces single-frame force
RMSE from 2.799\,N to 2.321\,N, a relative reduction of 17.1 \%,
and improves cross-sensor retrieval R@1 from 21.27\% to 34.38\%,
a gain of 13.11 percentage points, compared with the strongest
baseline for each metric.
For downstream manipulation, TaF-Policy achieves an average
success rate of 51.3\% across three UniVTAC
simulation~\cite{chen2026univtac} tasks, outperforming the
strongest evaluated baseline at 47.3\% by 4.0 percentage points.
On the three real-world tasks evaluated by all methods,
TaF-Policy achieves an average success rate of 40.0\%,
compared with 18.3\% for the strongest baseline,
a gain of 21.7 percentage points.
These results support the utility of wrench-grounded tactile
representations for both cross-sensor perception and
contact-rich manipulation.

Our main contributions are as follows:
\begin{itemize}
    \item \textbf{Large-scale tactile--wrench data infrastructure.}
    We develop a dedicated acquisition platform and collect over 100 million tactile--wrench pairs from 24 sensors spanning four sensing families.

    \item \textbf{Wrench-grounded unified tactile representations.}
    We introduce a unified tactile encoder with sensor-specific input adaptation and shared encoding. Wrench reconstruction and alignment with targets from a momentum-updated wrench encoder ground the representations in measured forces and torques.

    \item \textbf{Cross-sensor benchmark and manipulation policy.}
    We establish a benchmark for heterogeneous tactile sensors and demonstrate improvements in wrench estimation and cross-sensor retrieval. We also introduce TaF-Policy and evaluate its use of the learned representations in simulated and real-world manipulation tasks.
\end{itemize}

\section{Related Work}
\label{sec:related_work}

\paragraph{Heterogeneous tactile sensing.}
Tactile sensors are inherently heterogeneous because they rely on different sensing principles.
Optical tactile sensors produce image sequences~\cite{yuan2017gelsight,wu2025humanft,zhang2023gelstereo,lin20239dtact}, while magnetic and resistive sensors provide multichannel signals and pressure maps~\cite{bhirangi2025anyskin,huang2026flexitac}.
Even among optical sensors, differences in elastomer properties and optical layouts alter contact appearance~\cite{yuan2017gelsight}.
Fabrication variability can also cause distribution shifts between instances of the same sensor~\cite{bhirangi2025anyskin}.
This heterogeneity hinders data reuse and cross-sensor transfer, limiting scalable tactile representation learning.
We address this challenge using measured six-dimensional wrenches as a shared physical reference, with consistent force and torque definitions under common reference conventions.

\begin{table*}[t]
\centering
\caption{Comparison of tactile datasets paired with force or wrench measurements.}
\label{tab:datasets}
\begin{tabular}{lcccc}
\toprule
Dataset & Modality (Types) & Sensor Types & Contact Diversity & Paired Observations \\
\midrule
Sparsh Force\cite{higuera2024sparsh} 
& Vision tactile (1) & 2 & 3 indenters & $1.5 \times 10^5$ pairs \\
RCT\cite{he2026rct} 
& Vision tactile (1) & 1 & 122 materials & $2.9\times10^4$ pairs \\
TaF\cite{huang2026taf} 
& Vision/Pressure tactile (2) & 6 & 408 probe--sensor pairs& $10^7$ pairs \\
ToucHD-Force\cite{feng2026anytouch} 
& Vision tactile (1) & 5 & 71 indenters & $7.2\times10^5$ pairs \\
\textbf{Ours} 
& \textbf{Heterogeneous tactile (4)} 
& \textbf{24} 
& \textbf{1,445 probe--sensor pairs} 
& \textbf{$10^8$ pairs} \\
\bottomrule
\end{tabular}
\end{table*}

\paragraph{Tactile representation learning.}
Scaling tactile representation learning requires reusable features across
sensors and modalities. Existing approaches pursue this goal through
self-supervised pretraining, shared architectures, and representation
alignment. Sparsh~\cite{higuera2024sparsh} learns reusable features
through self-supervised pretraining, while T3~\cite{zhao2024transferable}
combines sensor-specific encoders with a shared Transformer trunk.
HTT~\cite{bi2026heterogeneous} and TactX~\cite{park2026tactx}
further align observations across sensors, with TactX spanning optical,
magnetic, and resistive modalities. Cross-modal methods extend alignment
to visual context~\cite{huang2026ht} and language:
CLTP~\cite{ma2026cltp} aligns tactile 3D point clouds with images and
contact-state descriptions, while FG-CLTP~\cite{ma2026fg} introduces
numerical tokens for fine-grained quantitative alignment.

Beyond observational alignment, physical supervision grounds tactile
representations in contact mechanics. AnyTouch~2~\cite{feng2026anytouch}
uses paired touch--force data to predict three-dimensional forces and
their changes. Closest to our work, UniForce~\cite{chen2026uniforce}
learns patch-wise latent force maps using quasi-static equilibrium
constraints and self- and cross-sensor reconstruction, without measured
force labels during pretraining. However, scalable physical supervision and a common representation
grounded in directly measured six-dimensional wrenches remain lacking.
In contrast, our framework enables autonomous large-scale data collection
and jointly learns tactile and wrench representations, using measured
wrenches as a common physical anchor for sensor-agnostic and physically
grounded representations.

\paragraph{Tactile learning for robotic manipulation.}
Tactile sensing complements vision with contact information,
supporting robust grasping and manipulation when visual
observations are limited or occluded~\cite{huang2026tactile,jiang2026caat,cheng2026tau}.
T-Rex~\cite{niu2026t} combines a temporal tactile encoder
with a variable-rate policy architecture to refine actions
using high-frequency tactile feedback.
Recent work also uses physical supervision to learn contact
representations relevant to action generation~\cite{helmut2026tactile,ma2026fg}.
For example, TaF-VLA~\cite{huang2026taf} learns separate
VQ-VAE~\cite{van2017neural} codebooks for synchronized wrench
and pressure-map sequences, aligns temporal tactile features
with wrench tokens through contrastive learning, and integrates
the tactile encoder into vision-language-action policies.
Similarly, NeoForce~\cite{team2026n_0} combines dense
three-axis force-field reconstruction with temporal latent
prediction for embodied manipulation.
While these methods connect tactile learning with manipulation,
our focus is on transferring representations pretrained across
heterogeneous sensors to downstream policies.
We evaluate this transfer using TaF-Policy, with the tactile
encoder frozen during policy training.

\begin{figure}[t]
    \centering
    \includegraphics[width=\linewidth]{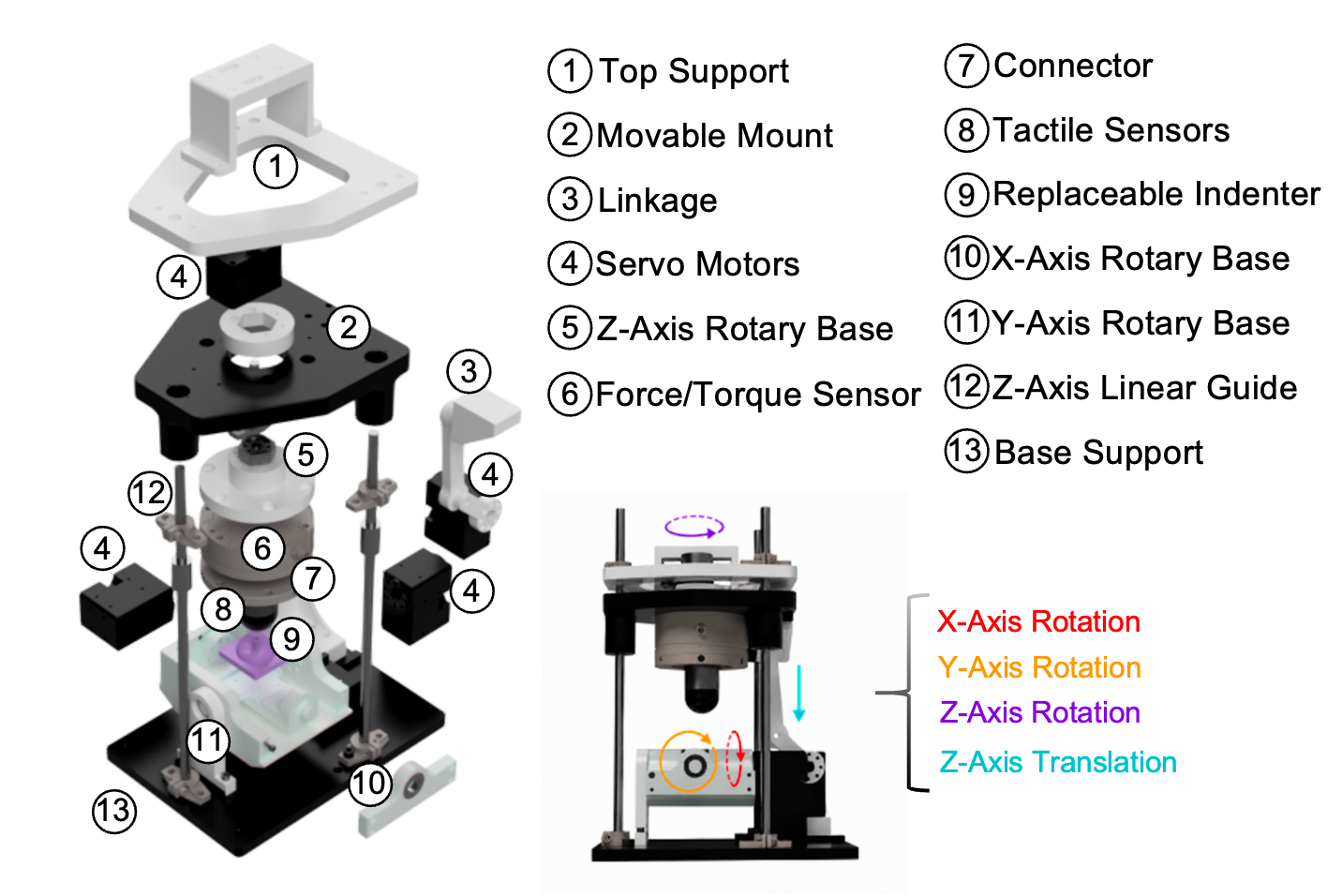}
    \caption{
        Acquisition apparatus for synchronized tactile--wrench
        recording, shown in exploded and assembled views.
    }
    \label{fig:dataset_setup}
\end{figure}
\section{Dataset Collection}
\label{sec:benchmark}

We collect paired tactile observations and six-dimensional wrench measurements using the platform in Fig.~\ref{fig:dataset_setup}.
A Dynamixel actuator controls the yaw rotation of the tactile and force/torque sensors, while the probe rotates about the $x$ and $y$ axes to vary contact orientation.
Pressing along the $z$ axis is controlled by adjusting the actuator torque.
Unlike TaF-VLA~\cite{huang2026taf}, which uses matched probes to collect tactile and force measurements in separate contacts under corresponding motions, our platform records both signals concurrently during the same physical contact, avoiding mismatches between separate contact events.
To ensure data diversity, the platform supports heterogeneous tactile sensors, interchangeable contact objects, and multiple contact poses per probe.
Table~\ref{tab:datasets} compares sensor coverage, contact diversity, and paired observation counts across related tactile datasets.
Details of the acquisition platform, reference wrench measurements, and sensor inventory are provided in Appendices~A.1--A.3 of the supplementary material.

\begin{figure*}[t]
    \centering
    \includegraphics[width=\linewidth]{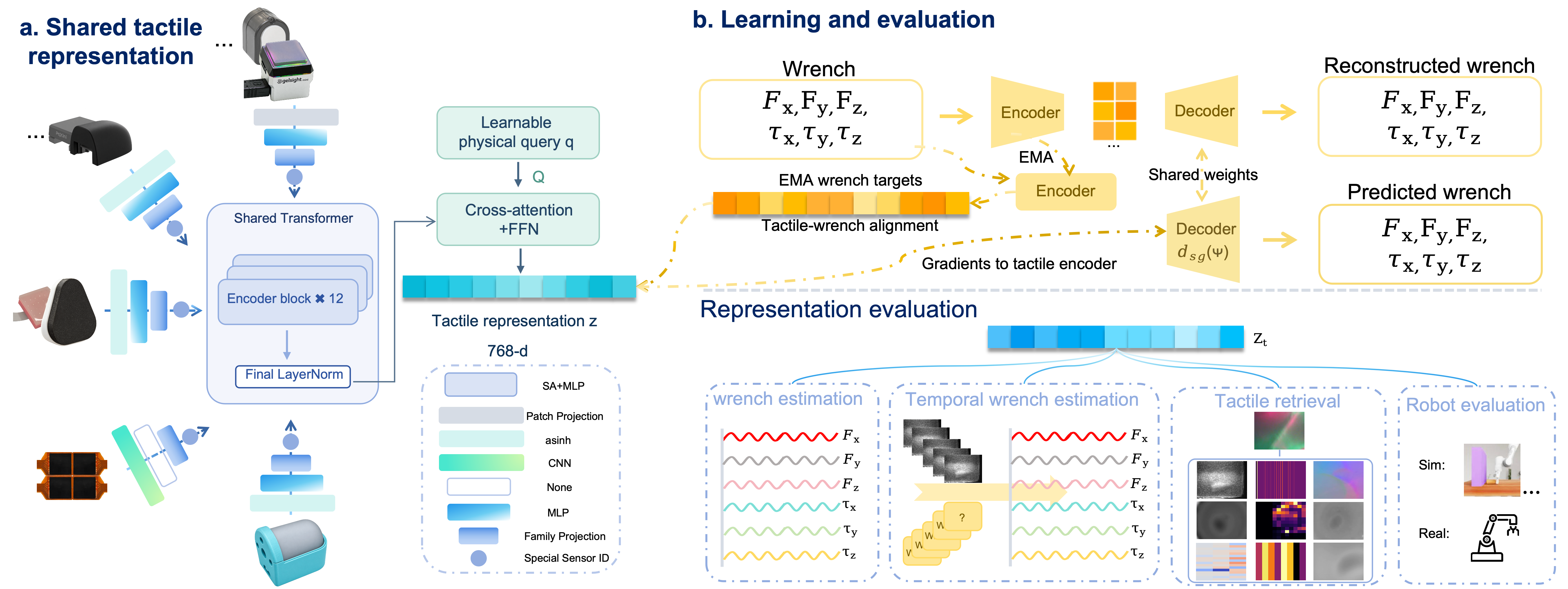}
    \caption{
        Overview of our methodology for wrench-grounded tactile representation learning and downstream evaluation.
        (a) Sensor-specific input adapters and family-specific projections map heterogeneous tactile observations into tokens for a shared Transformer.
        A learnable physical query aggregates the encoded tokens into a 768-dimensional continuous tactile representation.
        (b) Six-dimensional wrench reconstruction and alignment with EMA wrench embeddings provide shared physical supervision.
        The learned representations are evaluated on single-frame wrench estimation, temporal wrench estimation with historical wrench inputs, cross-sensor retrieval, and robotic manipulation using TaF-Policy in simulation and the real world.
    }

    \label{fig:pipeline}
\end{figure*}

\section{Tactile Representation Learning}
\label{sec:model}

Tactile sensors differ in observation format and contact response,
which pose challenges for shared representation learning.
To bridge the gap, we address this heterogeneity
through sensor-specific input adapters and shared physical supervision, as shown in Fig.~\ref{fig:pipeline}. Specifically, 
the adapters convert heterogeneous observations into tokens that are processed by a shared encoder. The encoded representation is then paired with wrench measurements, providing a common physical reference across sensors.
After training, the encoder produces tactile representations
for downstream tasks without wrench measurements at inference time.

\paragraph{Problem Formulation and Overview.}
We aim to learn wrench-grounded representations across heterogeneous tactile sensors.
Let $\mathcal{S}$ denote the sensor set and $\mathcal{D}$ a dataset of $N$ tactile--wrench pairs:
\begin{equation}
    \mathcal{D}
    =
    \left\{
        \left(x_i, \mathbf{w}_i, s_i\right)
    \right\}_{i=1}^{N},
\end{equation}
where $s_i \in \mathcal{S}$ identifies the sensor and
$x_i \in \mathcal{X}_{s_i}$ is its tactile observation.
The observation space $\mathcal{X}_{s_i}$ may vary across sensors.
Each observation is paired with a measured wrench
$\mathbf{w}_i =
[\mathbf{f}_i^{\top}, \boldsymbol{\tau}_i^{\top}]^{\top}
\in \mathbb{R}^{6}$,
where $\mathbf{f}_i \in \mathbb{R}^{3}$ and
$\boldsymbol{\tau}_i \in \mathbb{R}^{3}$ denote force and torque, respectively.

We learn a tactile encoder $E_{\theta}$ that maps these observations into a shared, continuous feature space:
\begin{equation}
    \mathbf{z}_i
    =
    E_{\theta}(x_i, s_i)
    \in \mathbb{R}^{d},
\end{equation}
where $\theta$ denotes the encoder parameters and $d$ the representation dimension.
We use paired wrench measurements to supervise the encoder for capturing interaction mechanics relevant to downstream tasks.
At inference, the encoder requires only tactile observations and sensor identity.

\paragraph{Unified Encoding of Heterogeneous Tactile Inputs.}
A key challenge in interfacing with diverse sensors is the variation in raw data formats.
We address this using input-specific tokenization to map tactile observations into a common 768-dimensional token space.
(1) For image-based sensors (visuotactile sensors), we concatenate the current image with its difference from a non-contact reference, then project the resulting six-channel input into patch tokens.
A sensor-specific residual MLP adapts each patch token:
\begin{equation}
    \widetilde{\mathbf{p}}_{i,j}
    = \mathbf{p}_{i,j}
    + \mathcal{A}_{s_i}(\mathbf{p}_{i,j}),
\end{equation}
where $\mathbf{p}_{i,j}$ denotes patch token $j$ of sample $i$, and $\mathcal{A}_{s_i}$ is the adapter for sensor $s_i$.
(2) For vector-valued observations, including magnetic and pressure-channel measurements, sensor-specific MLPs map inputs to tokens.
(3) For pressure-matrix observations, a convolutional frontend maps two-dimensional inputs to tokens.
All resulting tokens pass through family-specific projections before entering the shared encoder.

Next, we append a learnable sensor identity token and pass the sequence
through a shared, trainable Transformer initialized from
DINOv2~\cite{oquab2023dinov2}. A single learnable physical query
$\mathbf{q}$ aggregates the Transformer output $\mathbf{H}_i$ through
cross-attention and a feed-forward network, producing a pooled feature
$\mathbf{h}_i=\operatorname{QueryPool}(\mathbf{q},\mathbf{H}_i)$.
We then project $\mathbf{h}_i$ to the continuous tactile representation:
\begin{equation}
    \mathbf{z}_i
    = \tanh\!\left(
        W_z\operatorname{LN}(\mathbf{h}_i)+\mathbf{b}_z
    \right)
    \in \mathbb{R}^{768},
\end{equation}
where $\operatorname{LN}$ denotes layer normalization, and $W_z$ and
$\mathbf{b}_z$ are the learnable projection weight and bias.
Preprocessing, layer dimensions, and initialization details are provided
in Appendix~B of the supplementary material.

\paragraph{Wrench-Grounded Representation Learning.}
Paired wrench measurements provide a common physical reference for heterogeneous tactile observations.
Our goal is to learn a shared wrench representation space and align sensor-specific tactile embeddings with it.
We combine wrench reconstruction, which preserves force and torque information, with embedding alignment, which connects tactile features to this wrench-derived space.

Specifically, we encode the standardized wrench measurement $\bar{\mathbf{w}}_i$ as
$\mathbf{u}_i=g_{\phi}(\bar{\mathbf{w}}_i)\in\mathbb{R}^{768}$ via an encoder. Then,
a shared decoder $d_{\psi}$ reconstructs the wrench from both $\mathbf{u}_i$ and the previously encoded tactile sensor observation $\mathbf{z}_i$. Next,
an exponential moving average (EMA) copy of the wrench encoder $g_{\bar{\phi}}$, provides the alignment target
$\bar{\mathbf{u}}_i=g_{\bar{\phi}}(\bar{\mathbf{w}}_i)$.
We optimize:
\begin{equation}
    \begin{aligned}
    \mathcal{L}
    &=\operatorname{MSE}(d_{\psi}(\mathbf{u}),\bar{\mathbf{w}})\\
    &\quad+\operatorname{MSE}
      (d_{\operatorname{sg}(\psi)}(\mathbf{z}),\bar{\mathbf{w}})\\
    &\quad+\lambda\,\operatorname{MSE}
      (\mathbf{z},\operatorname{sg}(\bar{\mathbf{u}})),
    \end{aligned}
    \label{eq:wrench_objective}
\end{equation}
where MSE averages over samples and feature dimensions,
$\operatorname{sg}$ denotes stop-gradient, and $\lambda$ weights the alignment term.
The three terms serve the following roles, respectively:
(1) \textbf{Wrench reconstruction} trains the wrench encoder and decoder to reconstruct measured wrenches.
(2) \textbf{Tactile-to-wrench prediction} trains the tactile encoder to predict wrenches through the decoder, with gradients flowing to $\mathbf{z}$.
(3) \textbf{Representation alignment} aligns tactile embeddings with EMA wrench targets, which receive no gradients.

To evaluate representation quality, we freeze the pretrained tactile encoder and train new prediction heads on its output $\mathbf{z}$.
These heads are separate from the pretraining decoder $d_{\psi}$.
Details of the objective and EMA updates are provided in Appendix~B.

\paragraph{TaF-Policy: Wrench-Grounded Manipulation.}
We introduce \textbf{TaF-Policy}, a tactile-enhanced policy built on ACT~\cite{zhao2023learning}, to evaluate our pretrained tactile representations in manipulation.
The policy architecture is defined as follows.
At each policy query $t$, the pretrained tactile encoder processes observations from the left and right sensors using the input construction described above.
A shared, trainable linear projection maps each 768-dimensional representation to a tactile token:
\begin{equation}
    \mathbf{v}_t^{k}
    =
    W_{\mathrm{T}}E_{\theta}(x_t^{k},s_k)
    + \mathbf{b}_{\mathrm{T}}
    \in \mathbb{R}^{512},
    \qquad k\in\{\mathrm{L},\mathrm{R}\}.
\end{equation}
We concatenate the two tactile tokens with visual tokens from the head and wrist cameras.
Following ACT, we prepend the projected robot-state token and latent token.
Then, the Transformer encoder jointly processes the sequence, and the decoder predicts an action chunk $\hat{\mathbf{a}}_{t:t+H-1}$ of length $H$.

During policy training, we freeze the tactile encoder to
preserve its pretrained representations and prevent
action supervision from overwriting the learned tactile
features. We optimize the visual backbone, tactile projection,
and ACT policy using the standard action reconstruction and
latent regularization objectives of ACT.
During deployment, tactile features are recomputed at each
policy query without measured wrench inputs.

\section{Experiments}
\label{sec:experiments}

\begin{table*}[t]
\centering
\caption{
Benchmark results on the same 21 seen sensors spanning
four tactile sensor families.
F and T denote force and torque RMSE, respectively.
R@$K$ (\%) measures retrieval accuracy based on standardized 6D wrench distance, with candidates ranked by tactile-representation distance.
Results are reported as mean $\pm$ standard deviation over three seeds.
}
\label{tab:benchmark}
\footnotesize
\setlength{\tabcolsep}{2pt}
\renewcommand{\arraystretch}{1.15}
\begin{tabular*}{\textwidth}{
    @{\extracolsep{\fill}}lcccccc@{}
}
\toprule
& \multicolumn{2}{c}{Single-frame estimation}
& \multicolumn{2}{c}{Temporal estimation}
& \multicolumn{2}{c}{Cross-sensor retrieval} \\
\cmidrule(lr){2-3}
\cmidrule(lr){4-5}
\cmidrule(lr){6-7}
Method
& \shortstack{F RMSE\\(N) $\downarrow$}
& \shortstack{T RMSE\\(N$\cdot$m) $\downarrow$}
& \shortstack{F RMSE\\(N) $\downarrow$}
& \shortstack{T RMSE\\(N$\cdot$m) $\downarrow$}
& R@1 $\uparrow$
& R@5 $\uparrow$ \\
\midrule
DINOv2 ~\cite{oquab2023dinov2}
& $3.032 \pm 0.053$
& $0.0821 \pm 0.0015$
& $1.642 \pm 0.064$
& $0.0413 \pm 0.0003$
& $14.00 \pm 0.27$
& $40.25 \pm 2.56$ \\
Sparsh ~\cite{higuera2024sparsh}
& $2.886 \pm 0.155$
& $0.0788 \pm 0.0021$
& $1.579 \pm 0.016$
& $0.0364 \pm 0.0022$
& $14.21 \pm 0.59$
& $37.83 \pm 1.01$ \\
T3 ~\cite{zhao2024transferable}
& $2.820 \pm 0.131$
& $0.0780 \pm 0.0035$
& $1.482 \pm 0.019$
& $0.0372 \pm 0.0006$
& $17.38 \pm 1.05$
& $51.52 \pm 1.63$ \\
AnyTouch 2 ~\cite{feng2026anytouch}
& $2.799 \pm 0.203$
& $0.0775 \pm 0.0026$
& $1.515 \pm 0.040$
& $0.0387 \pm 0.0026$
& $21.27 \pm 0.80$
& $64.48 \pm 5.04$ \\
HTT ~\cite{bi2026heterogeneous}
& $2.971 \pm 0.024$
& $0.0783 \pm 0.0014$
& $1.438 \pm 0.094$
& $0.0343 \pm 0.0008$
& $14.78 \pm 0.46$
& $45.33 \pm 0.73$ \\
\midrule
\textbf{Ours ($\mathbf{z}$)}
& $\mathbf{2.321} \pm 0.084$
& $\mathbf{0.0657} \pm 0.0013$
& $\mathbf{1.237} \pm 0.068$
& $\mathbf{0.0322} \pm 0.0016$
& $\mathbf{34.38} \pm 1.35$
& $\mathbf{81.41} \pm 2.77$ \\
\bottomrule
\end{tabular*}
\end{table*}

\subsection{Experimental Setup}
\label{sec:exp_setup}

\paragraph{Data Splits.}
For data collected using the platform described in Sec.~\ref{sec:benchmark},
the training, validation, and test splits contain approximately
89.65\%, 2.45\%, and 7.90\% of the paired observations, respectively.
For few-shot evaluation, three of the 24 sensors are excluded
from source training.
Split details and adaptation protocols are provided in
Appendices~A.4 and C.4.

\paragraph{Comparisons Protocol.}
We evaluate our tactile encoder against
DINOv2~\cite{oquab2023dinov2},
Sparsh~\cite{higuera2024sparsh},
T3~\cite{zhao2024transferable},
AnyTouch~2~\cite{feng2026anytouch},
and HTT~\cite{bi2026heterogeneous}
on wrench estimation using the same 21 seen sensors spanning four sensing families.
We freeze each encoder and train a separate prediction head on its output features.
Optimization settings are provided in Appendix~B, and prediction-head architectures in Appendices~C.1 and C.2 of the supplementary material.
Table~\ref{tab:benchmark} reports core benchmark results as mean $\pm$ standard deviation across three random seeds.

\subsection{Benchmark on Quality of Representation}

\paragraph{Single-frame wrench estimation.}
Given a single tactile observation $x_t$ from sensor $s$, the task is to predict the corresponding wrench:
$\hat{\mathbf{w}}_t=f_{\mathrm{single}}(x_t,s)$.
As shown in Table~\ref{tab:benchmark}, our method achieves force and torque RMSEs of 2.321\,N and 0.0657\,N$\cdot$m, respectively, reducing errors by 17.1\% and 15.2\% relative to the strongest baseline for each metric.
These results show that our frozen tactile representations support more accurate force and torque estimation than the baselines.

\paragraph{Temporal wrench estimation.}
This task predicts the current wrench from five tactile observations sampled at one-second intervals and the four preceding measured wrenches:
$\hat{\mathbf{w}}_t=f_{\mathrm{temporal}}
(x_{t-4:t},\mathbf{w}_{t-4:t-1},s)$.
Our method achieves the lowest force and torque RMSEs of 1.237\,N and 0.0322\,N$\cdot$m, respectively, reducing errors by 14.0\% and 6.1\% relative to the strongest baseline for each metric.
The temporal input protocol and prediction-head architecture are detailed in Appendix~C.2 of the supplementary material.

\paragraph{Cross-sensor retrieval.}
\label{sec:retrieval_protocol}
Given a query observation $q$, we retrieve the candidate nearest in representation space:
$c^*=\arg\min_{c\in\mathcal{C}_q}
\|\mathbf{z}_q-\mathbf{z}_c\|_2$,
where $\mathbf{z}=E_\theta(x,s)$ for our model.
Each candidate set $\mathcal{C}_q$ contains 20 observations from other sensors, with balanced source sensors and identical sets across methods.
Ground-truth matches are the candidates nearest to the query in standardized six-dimensional wrench space:
\begin{equation}
    \begin{aligned}
    \bar{\mathbf{w}}&=(\mathbf{w}-\boldsymbol{\mu})
        \oslash\boldsymbol{\sigma},\\
    d_w(q,c)&=\left\|\bar{\mathbf{w}}_q
        -\bar{\mathbf{w}}_c\right\|_2,
    \end{aligned}
    \label{eq:retrieval_distance}
\end{equation}
where $\boldsymbol{\mu}$ and $\boldsymbol{\sigma}$ are the component-wise training-set mean and standard deviation, and $\oslash$ denotes elementwise division.
Our method achieves R@1 of 34.38\% and R@5 of 81.41\%\footnote{R@1 and R@5 denote the percentages of queries whose ground-truth match appears among the top 1 and top 5 retrieved candidates, respectively.}, exceeding the strongest baseline for each metric by 13.11 and 16.93 percentage points, respectively.

\subsection{Few-Shot Adaptation to Unseen Sensors}
\label{sec:fewshot}
\begin{figure}[t]
    \centering
    \includegraphics[width=\linewidth]{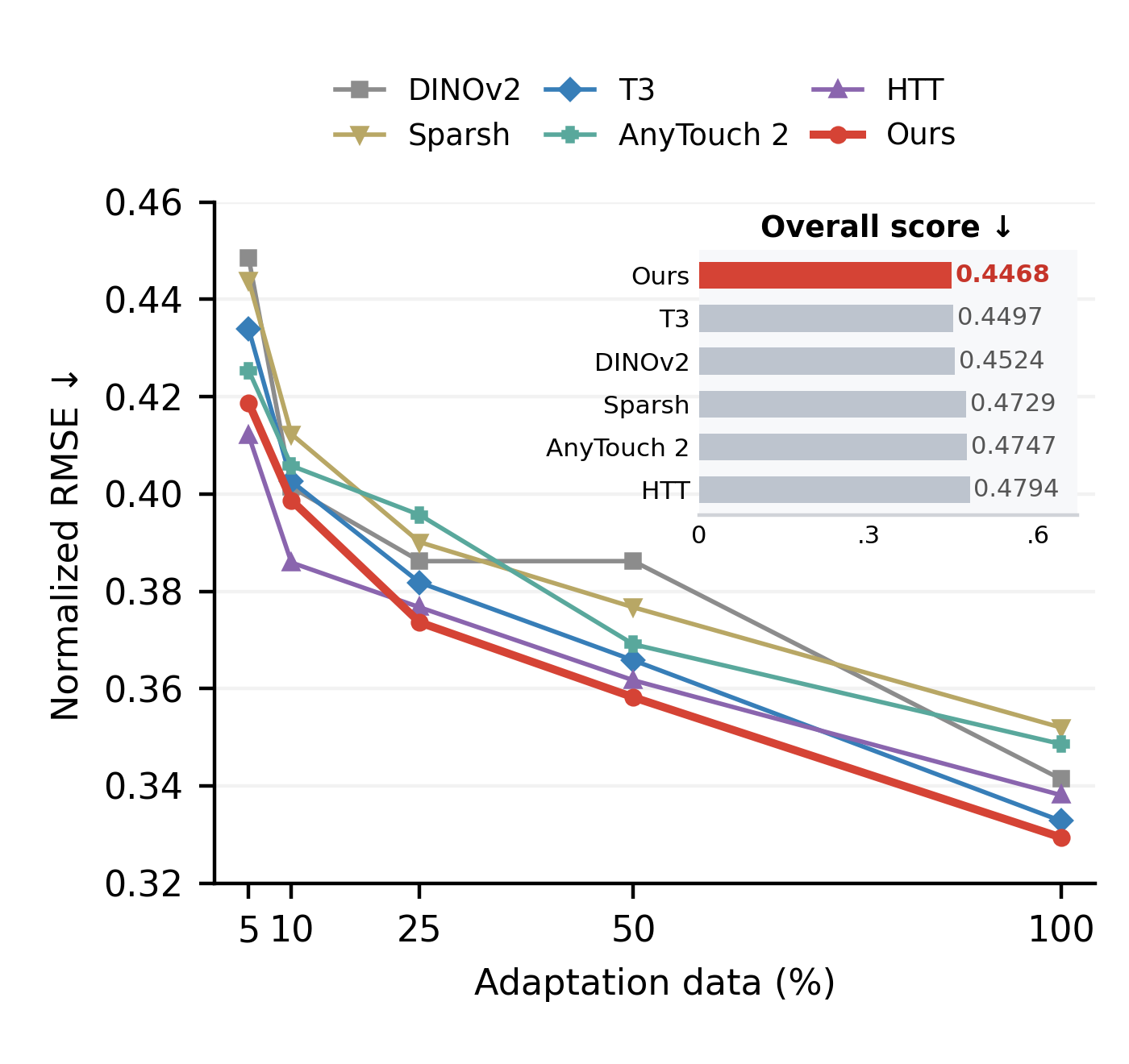}
    \caption{
        Few-shot adaptation on three unseen tactile sensors.
        Curves show temporal joint-adaptation results at different
        data budgets; the inset reports the hierarchical overall
        score across adaptation settings. Details are provided in
        Appendix~C.4.
    }
    \label{fig:fewshot}
\end{figure}

We hold out three of the 24 sensors during representation training.
For each unseen sensor, we use tactile--wrench pairs collected with two objects for adaptation, one other object for validation, and five additional objects for testing.
We jointly adapt the three sensors by training their sensor-specific input modules and a shared wrench prediction head while keeping the pretrained backbone frozen.
To assess how performance varies with data availability, we use 5--100\% of the labeled frames from each adaptation object.
Single-frame and temporal models are adapted separately, using identical object splits and data subsets across methods.
As shown in Fig.~\ref{fig:fewshot}, our method achieves the lowest overall score of 0.4468, computed by hierarchically aggregating normalized RMSE across adaptation settings (Appendix~B.3).
For temporal joint adaptation, it achieves the lowest errors when using 25\%, 50\%, and 100\% of the adaptation data.

\begin{table*}[t]
\centering
\caption{
    Ablation and design comparisons on the core benchmark. F and T denote force and torque RMSE; $D_1$ is the standardized 6D wrench distance of the top-ranked retrieval. Results are from a single random seed.
}
\label{tab:ablation}
\small
\setlength{\tabcolsep}{3pt}
\renewcommand{\arraystretch}{1.15}
\begin{tabular*}{\textwidth}{
    @{\extracolsep{\fill}}lccccccc@{}
}
\toprule
& \multicolumn{2}{c}{Single-frame estimation}
& \multicolumn{2}{c}{Temporal estimation}
& \multicolumn{3}{c}{Cross-sensor retrieval} \\
\cmidrule(lr){2-3}
\cmidrule(lr){4-5}
\cmidrule(lr){6-8}
Variant
& \shortstack{F RMSE\\(N) $\downarrow$}
& \shortstack{T RMSE\\(N$\cdot$m) $\downarrow$}
& \shortstack{F RMSE\\(N) $\downarrow$}
& \shortstack{T RMSE\\(N$\cdot$m) $\downarrow$}
& R@1 $\uparrow$
& R@5 $\uparrow$
& $D_1$ $\downarrow$ \\
\midrule
\textbf{Full model}
& \textbf{2.415}
& \textbf{0.0648}
& \textbf{1.242}
& \textbf{0.0318}
& 33.52
& \textbf{78.90}
& \textbf{1.581} \\
Direct wrench supervision
& 2.494
& 0.0684
& 1.385
& 0.0379
& 29.95
& 73.14
& 1.648 \\
w/o Alignment
& 2.481
& 0.0685
& 1.310
& 0.0370
& 28.00
& 76.10
& 1.652 \\
w/o Sensor-specific MLPs
& 3.291
& 0.0884
& 1.490
& 0.0384
& 25.90
& 67.81
& 1.761 \\
w/o Sensor Identity Token
& 2.628
& 0.0678
& 1.291
& 0.0323
& \textbf{33.90}
& 78.10
& 1.592 \\
\bottomrule
\end{tabular*}
\end{table*}

\subsection{Ablation and Design Analysis}
\label{sec:ablation}

To assess wrench-grounded learning and sensor-specific adaptation,
Table~\ref{tab:ablation} compares the full model with direct wrench
supervision and ablates the alignment loss, sensor-specific MLPs, and
sensor identity tokens. The full model improves all reported metrics
over direct supervision, reducing temporal force RMSE from 1.385\,N to
1.242\,N and increasing retrieval R@1 from 29.95\% to 33.52\%.
Removing the alignment loss also degrades all metrics. Removing
sensor-specific MLPs causes the largest degradation, increasing
single-frame force RMSE from 2.415\,N to 3.291\,N and reducing R@1
from 33.52\% to 25.90\%. Sensor identity tokens improve wrench
estimation and R@5, although R@1 is slightly higher without them
(33.90\% versus 33.52\%). Details are in Appendix~D.

\begin{figure}[t]
    \centering
    \includegraphics[width=\linewidth]{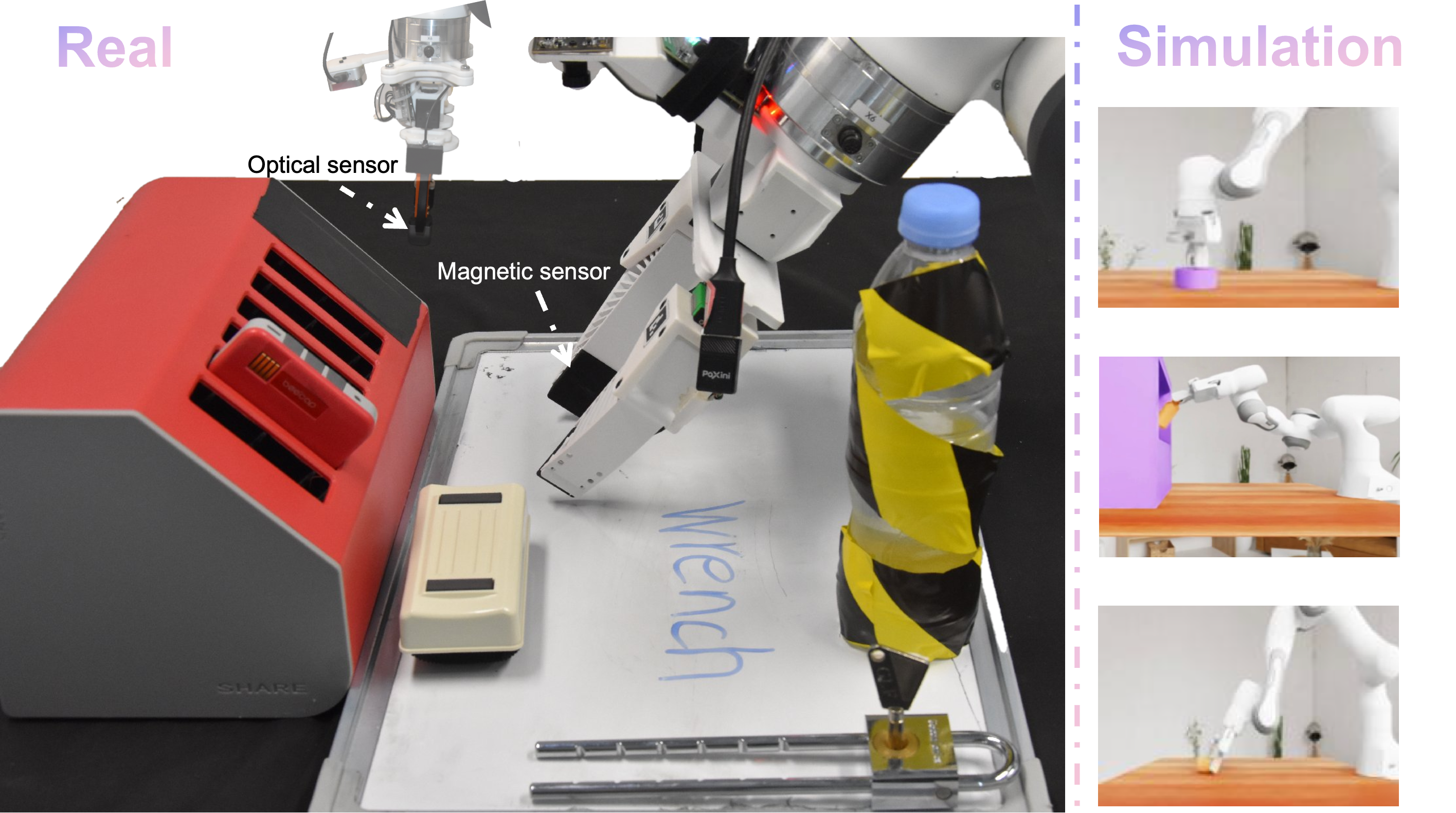}
    \caption{
        Real-world manipulation setups and UniVTAC simulation scenes. Four real-world tasks are presented: bottle lifting, power bank removal, key insertion, and board cleaning. The first three real-world tasks use Daimon tactile sensors, while the last task uses a PaXini tactile sensor. Three simulation tasks are presented: can lifting, key extraction, and bottle placement on a shelf.
    }
    \label{fig:robot_setup}
\end{figure}

\subsection{Robot Manipulation}
\label{sec:robot}

To assess the utility of the pretrained encoder and its learned representations, we evaluate TaF-Policy on simulated and real-world contact-rich manipulation tasks using the setups in Fig.~\ref{fig:robot_setup}.
Policy learning rates and batch sizes are provided in Table~S3 of the supplementary material.

\paragraph{Simulation Experiments on UniVTAC.}
We evaluate three UniVTAC~\cite{chen2026univtac} tasks:
Lift Can, Pull Out Key, and Put Bottle in Shelf.
Baselines include the official UniVTAC policy and policies
using T3 and AnyTouch~2 representations.
All methods use a common policy backbone and matched training
budgets, with 50 demonstrations and 50 evaluation episodes
per task. Success follows UniVTAC's task-specific criteria.
TaF-Policy achieves the highest success rate on all three
tasks, averaging 51.33\% versus 47.33\% for the strongest
baseline (Table~\ref{tab:robot_sim}).

\begin{table}[t]
    \centering
    \caption{
        Task success rates (\%) on three UniVTAC simulation tasks.
    }
    \label{tab:robot_sim}
    \small
    \setlength{\tabcolsep}{4pt}
    \renewcommand{\arraystretch}{1.15}
    \begin{tabular*}{\linewidth}{
        @{\extracolsep{\fill}}lcccc@{}
    }
        \toprule
        Method
        & \shortstack{Lift\\Can}
        & \shortstack{Pull Out\\Key}
        & \shortstack{Put Bottle\\in Shelf}
        & Avg. $\uparrow$ \\
        \midrule
        UniVTAC ~\cite{chen2026univtac}
        & 40 & 64 & 38 & 47.33 \\
        T3 ~\cite{zhao2024transferable}
        & 30 & 62 & 32 & 41.33 \\
        AnyTouch2 ~\cite{feng2026anytouch}
        & 32 & 68 & 28 & 42.67 \\
        \midrule
        \textbf{TaF-Policy (Ours)}
        & \textbf{44} & \textbf{70} & \textbf{40} & \textbf{51.33} \\
        \bottomrule
    \end{tabular*}
\end{table}

\paragraph{Real-World Experiments.}
We evaluate four tasks on a Franka arm: Lift Bottle, Remove Power Bank, and Key Insertion using DM-WS sensors, and Board Clean using a magnetic tactile sensor (PX6AX-GEN4-GR-M3525-Gripper).
Task definitions and success criteria are provided in
Appendix~C.5 of the supplementary material.
Each task uses 100 demonstrations and 20 evaluation trials per method under the same execution protocol.
We use the same baselines as in simulation, except that UniVTAC is evaluated only on the first three tasks because UniVTAC does not support magnetic-based tactile sensors.

For Board Clean, we adapt the sensor-specific frontend using 100\% of the adaptation data.
Adaptation updates the sensor-specific input module and wrench prediction head while keeping the backbone frozen.
The adapted tactile encoder remains frozen during policy training.

TaF-Policy achieves the highest success rate on all four tasks
(Table~\ref{tab:robot_real}), averaging 46.25\% versus 23.75\%
for the strongest baseline evaluated on all four tasks.
On the three tasks evaluated by all methods, it achieves
40.0\% compared with 18.3\% for the strongest baseline.

\paragraph{Analysis of Simulation and Real-World Results.}
We compare performance gains in simulation and the real world to examine how tactile observations may affect policy performance.
TaF-Policy improves average success rates over the strongest baselines by 4.0 percentage points in simulation and 22.5 percentage points in the real world.
We believe one possible reason is that simulated tactile images primarily capture contact geometry, whereas real observations contain richer force-dependent deformation cues.
Wrench-grounded representations may help the policy exploit these cues, contributing to the larger gains observed on real-world tasks.

\begin{table}[t]
    \centering
    \caption{
        Task success rates (\%) on four real-world manipulation tasks,
        evaluated over 20 trials per task. Board Clean uses a
        magnetic-based tactile sensor. UniVTAC is not evaluated on
        Board Clean because its encoder expects visuo-tactile images.
        Its average is computed over the first three tasks; averages
        for the other methods are computed over all four tasks.
    }
    \label{tab:robot_real}
    \footnotesize
    \setlength{\tabcolsep}{2pt}
    \renewcommand{\arraystretch}{1.15}
    \begin{tabular*}{\linewidth}{
        @{\extracolsep{\fill}}lccccc@{}
    }
        \toprule
        Method
        & \shortstack{Lift\\Bottle}
        & \shortstack{Remove\\Power Bank}
        & \shortstack{Key\\Insert}
        & \shortstack{Board\\Clean}
        & Avg. $\uparrow$ \\
        \midrule
        UniVTAC
        & 15 & 20 & 10 & -- & 15.00 \\
        T3
        & 20 & 30 & 5 & 35 & 22.50 \\
        AnyTouch2
        & 20 & 25 & 10 & 40 & 23.75 \\
        \midrule
        \textbf{TaF-Policy (Ours)}
        & \textbf{35} & \textbf{60} & \textbf{25}
        & \textbf{65} & \textbf{46.25} \\
        \bottomrule
    \end{tabular*}
\end{table}

\section{Conclusion and Limitations}
\label{sec:conclusion}
In this paper, we present a dataset of over 100 million
tactile--wrench pairs from 24 sensors and a framework that uses
wrenches as a common reference for learning across heterogeneous
sensors. The resulting representations outperform the evaluated
baselines on wrench estimation and cross-sensor retrieval, achieve
the best overall score in few-shot adaptation, and support
TaF-Policy, which achieves the highest average success rates in
both simulation and real-world manipulation.

\paragraph{Limitations and failure cases.}
Adapting to an unseen sensor still requires labeled tactile--wrench
pairs and training a sensor-specific input module, limiting deployment
when such data are difficult to collect. In real-world manipulation,
the policy also remains unreliable on tasks requiring precise alignment
or stable grasping: Key Insertion can fail due to key--keyhole
misalignment, while Lift Bottle can fail due to grasp instability and
slippage. Reducing the need for sensor-specific labeled data and
improving closed-loop manipulation remain directions for future work.

{\small

}

\clearpage
\appendix
\setcounter{figure}{0}
\setcounter{table}{0}
\setcounter{equation}{0}
\renewcommand{\thefigure}{S\arabic{figure}}
\renewcommand{\thetable}{S\arabic{table}}
\renewcommand{\theequation}{S\arabic{equation}}
\renewcommand{\theHfigure}{supp.\arabic{figure}}
\renewcommand{\theHtable}{supp.\arabic{table}}
\renewcommand{\theHequation}{supp.\arabic{equation}}
\pdfbookmark[0]{Supplementary Material}{supplementary}
\twocolumn[{\centering\Large\bfseries Supplementary Material\par\vspace{1em}}]
This supplement provides details on data collection, implementation, and
representation evaluation. Appendix~\ref{sec:supp_data} describes
the acquisition system, dataset coverage, and splits.
Appendix~\ref{sec:supp_training} provides shared implementation and evaluation
details. Appendix~\ref{sec:supp_evaluation} defines evaluation protocols,
and Appendix~\ref{sec:supp_ablation} specifies the ablation
and design comparison settings.

\section{Data Acquisition and Dataset Details}
\label{sec:supp_data}

\subsection{Acquisition Hardware and Procedures}
\label{sec:supp_acquisition}

Our acquisition system records tactile observations together with a six-axis
wrench reference comprising three force and three torque components.
Figure~\ref{fig:dataset_setup}
shows the mounting arrangement and measurement chain. We use a Dynamixel actuator to apply a downward load. Data collection includes pressing, sliding, rotation, and rolling. Across the 101,298,854 recorded frames, the observed force
magnitude ranges from 0 to 90.81\,N, and the torque magnitude
ranges from 0 to 1.67\,N$\cdot$m.
These magnitudes are computed as the per-frame Euclidean norms
of the three force and three torque components, respectively,
using labels before training standardization.
The ranges describe observed dataset values, rather than
the sensor's rated measurement range.

\subsection{Reference Wrench Measurements}
\label{sec:supp_quality}

We use an ATI Axia80-ZC22 six-axis force/torque sensor to
record reference wrench measurements. Measurements are
expressed in the sensor's native coordinate frame, with
the positive $F_z$ direction opposite to gravity. Forces
and torques are recorded in N and N$\cdot$m, respectively,
with torques referenced to the sensor origin. Bias correction
is performed before each data collection run.

\subsection{Sensor Inventory and Interaction Coverage}
\label{sec:supp_inventory}

Table S1 lists 24 tactile sensor entries across four families:
optical-based, magnetic-based, pressure-based, and
piezoresistive pressure-matrix sensors. Here, sensor identities are
distinguished by sensing model and working principles. Visuotactile sensors include those with and without markers, whereas Magnetic-based
sensors include both magnetic-field and force-valued inputs. 
Data collection includes pressing, sliding, rotation, and rolling.
The dataset includes more than 200 distinct probe geometries. It contains 1,445 probe--sensor pairs, each denoting a distinct combination of one probe and one sensor.

The dataset inventory totals 101,298,854 frames and 30,503 recorded
episodes across the training and evaluation directories,
including the three held-out sensors described in the main manuscript. Each saved data row counts as one frame, including
its paired tactile observation and wrench label.

\begin{table*}[t]
\centering
\caption{Sensor inventory and recorded data volume.
Input shapes describe saved observations before model preprocessing:
$H\times W\times C$ for images, channel counts for vectors,
and array dimensions for pressure matrices.
$\dagger$ marks the three sensors held out from source training.}
\label{tab:supp_inventory}
\footnotesize
\setlength{\tabcolsep}{4pt}
\renewcommand{\arraystretch}{1.08}
\begin{tabular*}{\textwidth}{@{\extracolsep{\fill}}llcrr@{}}
\toprule
Sensor & Family & Input shape & Episodes & Frames \\
\midrule
9DTact\cite{lin20239dtact}
& Optical-based & $240\times320\times3$
& 2,366 & 6,555,482 \\
DM-Tac WL
& Optical-based & $240\times320\times3$
& 2,098 & 6,190,543 \\
DM-Tac WS
& Optical-based & $240\times320\times3$
& 849 & 3,126,433 \\
NLiPsTac\cite{qin2026nlipscalib}
& Optical-based & $240\times320\times3$
& 1,640 & 4,881,827 \\
PPTac type 1\cite{lin2025pp}
& Optical-based & $240\times320\times3$
& 922 & 3,755,271 \\
PPTac type 2\cite{lin2025pp}
& Optical-based & $240\times320\times3$
& 2,088 & 6,236,062 \\
PPTac type 3\cite{lin2025pp}
& Optical-based & $240\times320\times3$
& 1,427 & 4,245,576 \\
Custom-designed ($6\times6$ markers)
& Optical-based & $380\times380\times3$
& 858 & 1,809,205 \\
Custom-designed ($7\times7$ markers)
& Optical-based & $380\times380\times3$
& 131 & 231,435 \\
Custom-designed ($8\times8$ markers)
& Optical-based & $380\times380\times3$
& 343 & 798,315 \\
Custom-designed (without markers)
& Optical-based & $380\times380\times3$
& 847 & 1,775,553 \\
DIGIT\cite{lambeta2020digit}
& Optical-based & $240\times320\times3$
& 886 & 2,604,158 \\
GelSight Mini (with markers)\cite{yuan2017gelsight}
& Optical-based & $240\times320\times3$
& 1,445 & 3,020,400 \\
GelSight Mini (without markers)\cite{yuan2017gelsight}
& Optical-based & $240\times320\times3$
& 974 & 2,184,516 \\
MiniPPTac type 1$^{\dagger}$
& Optical-based & $240\times320\times3$
& 8 & 360,000 \\
MiniPPTac type 2$^{\dagger}$
& Optical-based & $240\times320\times3$
& 8 & 324,000 \\
\midrule
AnySkin\cite{bhirangi2025anyskin}
& Magnetic-based(Magnetic-input) & 15
& 2,871 & 16,165,846 \\
SuperMag\cite{hou2025supermag}
& Magnetic-based(Magnetic-input) & 64
& 406 & 2,324,372 \\
\midrule
PX6AX-GEN3-DP-S2015-Elite
& Magnetic-based(Force-input)& 159
& 804 & 4,762,455 \\
PX6AX-GEN3-IP-S1610-Elite
& Magnetic-based(Force-input) & 78
& 1,031 & 6,008,741 \\
PX6AX-GEN3-MC-M2020-Elite
& Magnetic-based(Force-input) & 30
& 905 & 5,270,261 \\
PX6AX-GEN4-GR-M3525-Gripper$^{\dagger}$
& Magnetic-based(Force-input) & 405
& 24 & 144,000 \\
\midrule
TwinTac\cite{huang2025twintac}
& pressure-based & 8
& 2,972 & 8,700,502 \\
\midrule
Pressure matrix sheet
& Piezoresistive-based & $12\times12$
& 4,600 & 9,823,901 \\
\midrule
\textbf{Total (24 sensor entries)}
& -- & --
& \textbf{30,503} & \textbf{101,298,854} \\
\bottomrule
\end{tabular*}

\smallskip
\begin{minipage}{\textwidth}
\footnotesize
Counts aggregate the full recorded inventory across the
training and evaluation directories, including the three
held-out sensors. Few-shot evaluation uses reduced subsets
of this inventory, as detailed in Appendix~C.4.
\end{minipage}
\end{table*}

\subsection{Dataset Splits}
\label{sec:supp_protocol}

\paragraph{Object and episode separation.}
The main benchmark uses seen hardware while holding out indenter objects and episodes for evaluation. The train, validation, and test partitions contain approximately 89.65\%, 2.45\%, and 7.90\% of paired observations, respectively.

\paragraph{Unseen-sensor separation.}
We hold out three sensors from source representation
learning and source prediction-head training: Minpptac\_type1, Minpptac\_type2,
and PX6AX-GEN4-GR-M3525-Gripper. The remaining
21 sensors form the source set for training. The data of each held-out sensor provides
two objects for labeled adaptation, one for validation, and
five for testing. 
Adaptation settings are specified in Appendix~\ref{sec:supp_ood}. Table~\ref{tab:supp_splits} further summarizes the number of objects
assigned to each partition for the held-out sensors.

\begin{table}[t]
\centering
\caption{Indenter object splits for adaptation to the three held-out
sensors. Each sensor uses disjoint object sets for adaptation,
validation, and testing.}
\label{tab:supp_splits}
\small
\begin{tabular}{lc}
\toprule
Partition & Objects per held-out sensor \\
\midrule
Adaptation & 2 \\
Validation & 1 \\
Test & 5 \\
\bottomrule
\end{tabular}
\end{table}

\section{Implementation and Evaluation Details}
\label{sec:supp_training}

\subsection{Representation Learning and Optimization}
\label{sec:supp_optimization}

Table~\ref{tab:supp_training} summarizes the optimization
settings for each training stage. 

\paragraph{Tactile input architecture.}
For image-based sensors, the input concatenates the current RGB image
$I_i$ with its difference from a non-contact reference image,
$I_i-I_i^0$.
The RGB channels are normalized, and the resulting six-channel input is resized to $238\times322$ and
partitioned into $14\times14$ patches, yielding $17\times23=391$ tokens.
The six-channel patch projection is initialized by duplicating the
pretrained three-channel weights and scaling them by $1/2$.
Each image sensor is equipped with a residual adapter comprising
layer normalization, a Linear layer ($768\rightarrow32$), GELU,
and a second Linear layer ($32\rightarrow768$).
The weights and biases of the final layer are initialized to zero,
ensuring that the adapter initially preserves the original patch tokens.
Interpolated pretrained positional embeddings are added after adaptation.

For vector-valued sensors with input width $d_s$, an elementwise
$\operatorname{asinh}$ transformation is applied to compress the
signal dynamic range. Then, we process it via MLP,
with dimensions $d_s\rightarrow256\rightarrow32\times768$ and an
intermediate GELU activation. The output is reshaped into 32 tokens,
to which learnable positional embeddings are added.

For pressure-matrix observations, we converted them into single-channel
floating-point matrices at their native spatial resolution. The frontend
consists of three convolutional layers with channel widths
$1\rightarrow64\rightarrow128\rightarrow768$, kernel sizes $5,3,3$,
stride 2, and padding $2,1,1$, respectively. Each convolution is
followed by GELU. The resulting feature map is flattened along its
spatial dimensions to form 768-dimensional tokens.

All tokens have a width of 768 and pass through a family-specific
$768\rightarrow768$ linear projection initialized to the identity.
A learnable sensor identity token is then appended before the shared
backbone. 

The backbone retains the DINOv2 ViT-B/14 CLS token and its
positional embedding, followed by 12 Transformer blocks with 12
attention heads each and a final layer normalization. All backbone
parameters remain trainable during representation learning.
A learnable query aggregates the backbone outputs through
pre-normalized cross-attention, followed by a residual feed-forward
network with dimensions $768\rightarrow3072\rightarrow768$ and
GELU activation. Finally, layer normalization, a $768\rightarrow768$
linear projection, and $\tanh$ produce the representation
$\mathbf{z}$.

\paragraph{Wrench branch and objective.}
To provide physically grounded supervision for tactile representation
learning, we introduce a wrench branch that encodes and reconstructs
six-dimensional wrench measurements. Fixed component-wise means
$\boldsymbol{\mu}$ and standard deviations $\boldsymbol{\sigma}$ give
$\bar{\mathbf{w}}_i=(\mathbf{w}_i-\boldsymbol{\mu})\oslash
\boldsymbol{\sigma}$, where $\oslash$ denotes elementwise division.
The wrench encoder $g_{\phi}$ is an MLP
$6\rightarrow128\rightarrow128\rightarrow768$ with GELU between
linear layers and a final $\tanh$.
The decoder $d_{\psi}$ is an MLP
$768\rightarrow128\rightarrow128\rightarrow6$, also with intermediate
GELU activations and a linear output.

For a minibatch $\mathcal{B}$, let
$\mathbf{u}_i=g_{\phi}(\bar{\mathbf{w}}_i)$ and
$\bar{\mathbf{u}}_i=g_{\bar{\phi}}(\bar{\mathbf{w}}_i)$, where
$g_{\bar{\phi}}$ is the EMA encoder. The objective combines wrench
reconstruction, tactile-to-wrench prediction, and cross-modal
representation alignment. Specifically, wrench reconstruction trains the wrench encoder and
decoder, tactile-to-wrench prediction encourages the tactile
representation to predict measured wrenches, and alignment brings
the tactile representation closer to the EMA wrench representation. The losses are
\begin{equation}
\begin{aligned}
\mathcal{L}_{\mathrm{w}}
&=\frac{1}{6|\mathcal{B}|}\sum_{i\in\mathcal{B}}
  \|d_{\psi}(\mathbf{u}_i)-\bar{\mathbf{w}}_i\|_2^2,\\
\mathcal{L}_{\mathrm{t}}
&=\frac{1}{6|\mathcal{B}|}\sum_{i\in\mathcal{B}}
  \|d_{\operatorname{sg}(\psi)}(\mathbf{z}_i)
    -\bar{\mathbf{w}}_i\|_2^2,\\
\mathcal{L}_{\mathrm{align}}
&=\frac{1}{768|\mathcal{B}|}\sum_{i\in\mathcal{B}}
  \|\mathbf{z}_i-\operatorname{sg}(\bar{\mathbf{u}}_i)\|_2^2,\\
\mathcal{L}
&=\mathcal{L}_{\mathrm{w}}+\mathcal{L}_{\mathrm{t}}
  +0.2\,\mathcal{L}_{\mathrm{align}}.
\end{aligned}
\label{eq:supp_representation_objective}
\end{equation}
Here $\operatorname{sg}$ denotes stop-gradient. The factors 6 and 768
make each term a mean over both samples and output dimensions.

\paragraph{Gradient flow and EMA updates.}
The tactile encoder receives gradients from
$\mathcal{L}_{\mathrm{t}}$ and $\mathcal{L}_{\mathrm{align}}$,
which encourage physical wrench prediction and cross-modal alignment,
respectively.
In $\mathcal{L}_{\mathrm{t}}$, the decoder's parameters are detached,
but gradients are propagated through its input, allowing the tactile
representation to be optimized through the shared decoder.
Only $\mathcal{L}_{\mathrm{w}}$ updates the online wrench encoder and
decoder. 
The EMA encoder provides a slowly evolving target for representation
alignment. It receives no gradients and is initialized with
$\bar{\phi}_0=\phi_0$. After each optimizer step $k$, it is updated as
\begin{equation}
\begin{aligned}
\bar{\phi}_k&=\rho_k\bar{\phi}_{k-1}+(1-\rho_k)\phi_k,\\
\rho_k&=\begin{cases}
0.99,&1\leq k<1000,\\
0.999,&k\geq1000.
\end{cases}
\end{aligned}
\label{eq:supp_ema_update}
\end{equation}

\paragraph{Representation and prediction-head training.}
Our dataset comprises sequential tactile observations collected along
interaction trajectories. Training consists of two stages.

\textbf{Stage 1: Representation learning.}
We train the tactile encoder and online wrench branch using AdamW
with a global batch size of 256 and the fixed training budget specified
in Table S3. The tactile parameters use a learning rate of
$3\times10^{-5}$, while the online wrench encoder and decoder use
$10^{-3}$. Both parameter groups use a weight decay of 0.01, and
gradients are clipped to a global norm of 1.
Each global batch samples from all 21 seen sensors with approximately
equal numbers of frames per sensor. When the batch size is not divisible
by the number of sensors, the extra samples are rotated across sensors.
This sensor-balanced sampling ensures coverage of all seen sensors
without requiring a full pass through the dataset.

\textbf{Stage 2: Prediction head training.}
With the tactile encoder frozen, we independently train the single-frame
and temporal prediction heads for 10,000 steps each. Both use AdamW
with a batch size of 256, a learning rate of $10^{-3}$, a weight decay
of $10^{-4}$, and sensor-balanced sampling. The training objective is
MSE on standardized six-dimensional wrench measurements.

Both stages use constant learning rates and evaluate models every
1,000 steps. Representation checkpoints are selected by minimizing
sensor-balanced normalized RMSE, while prediction head checkpoints
are selected by minimizing sensor-balanced normalized MSE. These two
criteria induce the same ordering under the same aggregation.

\paragraph{Policy interface.}
For robot-policy training, the tactile encoder remains frozen, while
a shared trainable projection maps each sensor's 768-dimensional
continuous representation to a 512-dimensional ACT token.
The visual backbone, projection, and policy are jointly optimized
under matched data and training budgets to ensure a fair comparison.
The TaF-Policy architecture is detailed in the Tactile Representation
Learning section of the main paper. Each policy is evaluated over
50 episodes per task in simulation and 20 trials per task in the
real world.

\begin{table*}[t]
\centering
\caption{Training and adaptation settings.
All stages use AdamW. Batch sizes are global.
Weight decay is 0.01 for representation learning and
$10^{-4}$ for frozen probes and few-shot adaptation.
Few-shot schedules follow Appendix~\ref{sec:supp_ood};
robot-policy training steps vary by task.}
\label{tab:supp_training}
\footnotesize
\setlength{\tabcolsep}{4pt}
\renewcommand{\arraystretch}{1.15}

\begingroup
\newcommand{\traincell}[2]{%
  \parbox[t]{#1\textwidth}{\raggedright #2}}

\begin{tabular}{@{}lllll@{}}
\toprule
\traincell{0.18}{Stage}
& \traincell{0.17}{Learning rate}
& \traincell{0.07}{Batch}
& \traincell{0.24}{Training budget}
& \traincell{0.25}{Frozen modules} \\
\midrule

\traincell{0.18}{Representation learning}
& \traincell{0.17}{Tactile: $3\times10^{-5}$\\
                  Wrench: $10^{-3}$}
& 256
& \traincell{0.24}{100,000 steps}
& \traincell{0.25}{None in online branches; EMA updated separately} \\
\addlinespace[5pt]

\traincell{0.18}{Frozen probes}
& $10^{-3}$
& 256
& \traincell{0.24}{10,000 steps per head}
& \traincell{0.25}{Tactile encoder} \\
\addlinespace[5pt]

\traincell{0.18}{Few-shot adaptation}
& \traincell{0.17}{Inputs: $10^{-4}$\\
                  Head: $10^{-3}$}
& 256
& \traincell{0.24}{Single-frame: 20 equivalent epochs,
                   at most 5,000 steps\\
                   Temporal: 10 equivalent epochs,
                   at most 2,500 steps}
& \traincell{0.25}{Backbone; also the prediction head
                   for input-only adaptation} \\
\addlinespace[5pt]

\traincell{0.18}{Robot-policy training}
& $10^{-5}$
& 64
& \traincell{0.24}{Task-dependent steps; demonstrations per task:
                   50 in simulation, 100 in the real world}
& \traincell{0.25}{Tactile encoder} \\
\bottomrule
\end{tabular}
\endgroup
\end{table*}

\subsection{Baselines and Frozen Probes}
\label{sec:supp_baselines}

We evaluate tactile representations on the same set of 21 seen sensors.
For each sensor type not directly supported by a baseline, we introduce individual
learnable MLP adapters to map their observations to the input format
expected by that baseline. The adapters are trained through a wrench prediction head
using the mean squared error (MSE) between predicted and
component-wise standardized six-dimensional wrench measurements.
For each method, we select the checkpoint with the best validation
performance for evaluation. During frozen-probe evaluation, the
baseline encoders remain fixed, while a wrench prediction head is
trained on their output representations.

Our model produces a continuous tactile representation
$\mathbf{z}\in\mathbb{R}^{768}$.
Prediction-head architectures and task-specific evaluation protocols
are described in Appendix~\ref{sec:supp_evaluation}, and optimization
settings are provided in Appendix~\ref{sec:supp_optimization}.

\subsection{Evaluation Sets, Metrics, and Statistical Reporting}
\label{sec:supp_metrics}

\paragraph{Evaluation sets.}
The fixed single-frame evaluation set contains 567,000 frames
from 21 sensors. For each sensor, we select three objects,
three episodes per object, and 3,000 frames per episode,
yielding 189 episodes in total.

\paragraph{Error metrics.}
We report the following metrics, with equal weighting across sensors.

\textbf{(1) Force and torque RMSE.}
For sensor $s$ with $N_s$ samples, the force-component MSE is
\begin{equation}
M_s^F=\frac{1}{3N_s}\sum_{i=1}^{N_s}
\|\widehat F_{si}-F_{si}\|_2^2.
\end{equation}
The sensor-balanced force RMSE is
\begin{equation}
\mathrm{RMSE}_F=
\sqrt{\frac{1}{S}\sum_{s=1}^{S}M_s^F}.
\end{equation}
Torque RMSE is defined analogously using the three torque components.
These metrics retain their physical units of N and N$\cdot$m,
respectively.

\textbf{(2) Normalized MSE (NMSE).}
To account for differences in scale across wrench components, we
standardize each component using fixed source-training statistics.
For sensor $s$ with $N_s$ valid targets, the normalized MSE is
\begin{equation}
\mathrm{NMSE}_s
=\frac{1}{6N_s}\sum_{i=1}^{N_s}\sum_{j=1}^{6}
\left(
\frac{\hat{w}_{sij}-w_{sij}}{\sigma_j}
\right)^2,
\end{equation}
where $\sigma_j$ is the standard deviation of wrench component $j$
used for target standardization. The sensor-balanced NMSE is
\begin{equation}
\mathrm{NMSE}_{\mathrm{bal}}
=\frac{1}{S}\sum_{s=1}^{S}\mathrm{NMSE}_s.
\end{equation}

\textbf{(3) Normalized RMSE (NRMSE).}
For each sensor, NRMSE is the square root of its NMSE:
\begin{equation}
\mathrm{NRMSE}_s=\sqrt{\mathrm{NMSE}_s}.
\end{equation}
The sensor-balanced NRMSE is
\begin{equation}
\mathrm{NRMSE}_{\mathrm{bal}}
=\sqrt{\frac{1}{S}\sum_{s=1}^{S}\mathrm{NMSE}_s}.
\end{equation}
Both normalized metrics are dimensionless.

\paragraph{Few-shot aggregation.}
To summarize performance across few-shot adaptation settings, we
aggregate the normalized RMSE values from 12 conditions: single-frame
and temporal joint adaptation at five labeled-data proportions, and
single-frame and temporal input-only adaptation at 10\%. Each
condition's normalized RMSE is computed by averaging normalized MSE
equally across the three held-out sensors and then taking the square
root. Lower values indicate better performance.

We group the conditions into single-frame joint adaptation, temporal
joint adaptation, and input-only adaptation. For each method, we first
average the errors within each group and then average the three group
means with equal weight. This prevents a group from receiving more
weight solely because it contains more evaluated conditions.

Let $\mathcal{P}=\{5\%,10\%,25\%,50\%,100\%\}$ denote the labeled-data
proportions. For method $m$, let $e^{\mathrm{sf}}_{m,p}$ and
$e^{\mathrm{tmp}}_{m,p}$ denote the normalized RMSE values for
single-frame and temporal joint adaptation at proportion $p$.
Let $e^{\mathrm{in}}_{m,\mathrm{sf}}$ and
$e^{\mathrm{in}}_{m,\mathrm{tmp}}$ denote those for single-frame and
temporal input-only adaptation at 10\%. The group means are
\begin{equation}
\begin{aligned}
A_m &= \frac{1}{5}\sum_{p\in\mathcal{P}} e^{\mathrm{sf}}_{m,p},\\
B_m &= \frac{1}{5}\sum_{p\in\mathcal{P}} e^{\mathrm{tmp}}_{m,p},\\
C_m &= \frac{e^{\mathrm{in}}_{m,\mathrm{sf}}
             +e^{\mathrm{in}}_{m,\mathrm{tmp}}}{2}.
\end{aligned}
\end{equation}
The overall score is
\begin{equation}
S_m=\frac{A_m+B_m+C_m}{3}.
\label{eq:supp_hierarchical_score}
\end{equation}
This assigns a weight of $1/15$ to each joint-adaptation condition
and $1/6$ to each input-only condition. As a sensitivity check, we
also report the average that weights all 12 conditions equally.

\paragraph{Uncertainty.}
Results reported with $\pm$ in the main benchmark represent the mean
and standard deviation across three random seeds. Ablation experiments
and few-shot adaptation use a single random seed and are therefore
reported as single-run results. Few-shot adaptation details are provided
in Appendix~\ref{sec:supp_ood}.

\section{Evaluation Protocols and Metrics}
\label{sec:supp_evaluation}

This section specifies the evaluation protocols for the main-paper benchmarks and few-shot results. 

\subsection{Single-Frame Wrench Estimation}
\label{sec:supp_force}

We use an MLP probe to predict the current six-dimensional wrench from a frozen tactile representation. The probe consists of layer normalization, Linear $d\rightarrow256$, GELU, and Linear $256\rightarrow6$, where $d$ denotes the encoder feature dimension ($d=768$ for our model). The probe is trained independently of the wrench decoder used during representation learning, with the tactile encoder kept frozen throughout training. Predictions are made in standardized wrench coordinates and converted back to physical units for RMSE evaluation. All methods are evaluated on the same set of 21 seen sensors. Force and torque RMSE are reported separately and aggregated across sensors following Appendix~\ref{sec:supp_metrics}.

\subsection{Temporal Wrench Estimation}
\label{sec:supp_temporal}

The temporal probe uses tactile observations at offsets
$(-4,-3,-2,-1,0)$ seconds on the sampling clock.
The frozen encoder produces five representations
$\mathbf{z}_{t-4},\ldots,\mathbf{z}_t$. A single-layer GRU with hidden
dimension 256 processes the four historical pairs
$[\mathbf{z}_{t-j};\bar{\mathbf{w}}_{t-j}]$ for $j=4,3,2,1$ in
chronological order.
The final GRU hidden state is concatenated with the current representation
$\mathbf{z}_t$ and passed to an MLP with widths
$(256+d)\rightarrow256\rightarrow256\rightarrow6$ and intermediate
GELU activations, where $d=768$ for our model.

Historical wrench measurements are used exclusively by the temporal head,
while each tactile representation is computed solely from its corresponding
tactile observation and sensor identity. Neither the current target wrench
nor future observations are provided as inputs. All compared temporal
models use identical historical wrench inputs, target windows, and training
budgets, with the same head architecture except for input dimensions.
Optimization settings are provided in
Appendix~\ref{sec:supp_optimization}.

\subsection{Cross-Sensor Retrieval}
\label{sec:supp_retrieval}

We sample 100 query observations from each of the 21 seen sensors in
the fixed evaluation set described in Appendix~\ref{sec:supp_metrics},
yielding 2,100 retrieval trials. For each query, we uniformly sample
one candidate from each of the other 20 sensors. The candidate ranking
and ground-truth match follow the retrieval protocol described in the
main paper.

\subsection{Few-Shot Adaptation to Unseen Sensors}
\label{sec:supp_ood}

To evaluate adaptation to unseen sensors, we hold out three sensors
from representation pretraining: (1) MiniPPTac type 1 and type 2,
two optical tactile sensors, and (2) PX6AX-GEN4-GR-M3525-Gripper
(M3525), a magnetic tactile sensor. The following protocol describes
few-shot adaptation and evaluation on these three sensors.

We adapt the three sensors jointly using separate sensor-specific
input modules and a shared pretrained backbone. We evaluate two
settings: joint adaptation updates both the input modules and the
prediction head, whereas input-only adaptation updates only the
input modules. The backbone remains frozen in both settings.

\paragraph{Indenter objects and data splits.}
For each held-out sensor, we use data collected with two indenter
objects for adaptation, one for validation, and five for testing.
The three sets of objects are disjoint.

The adaptation pool consists of data from the two adaptation objects
for each sensor. It contains 36,000 frames from M3525 and 18,000
frames from each MiniPPTac sensor, for a total of 72,000 frames.

For joint adaptation, we use 5\%, 10\%, 25\%, 50\%, or 100\% of
the labeled frames from each adaptation object. These proportions
correspond to 3,600, 7,200, 18,000, 36,000, and 72,000 single-frame
samples in total, respectively. For input-only adaptation, we use 10\% of the labeled frames from
each adaptation object to train the sensor-specific input modules,
while keeping the backbone and prediction head frozen. Both settings are evaluated separately for single-frame and
temporal wrench estimation.

\paragraph{Sampling and temporal inputs.}
The nominal sampling rate is 60\,Hz for M3525 and 30\,Hz for both MiniPPTac sensors. For each sensor--object record, we divide the frames from each episode into five-second blocks. We shuffle the blocks within each episode and interleave blocks from different episodes to obtain a sampling order. Each data budget selects the corresponding percentage of frames from the beginning of this order. This produces nested subsets: frames selected at a smaller budget are also included at larger budgets.

For temporal adaptation, each input window contains five observations at offsets $(-4,-3,-2,-1,0)$ seconds relative to the prediction time. All five observations must belong to the same episode and be included in the selected subset. Measured wrenches are provided for the first four observations only.

\paragraph{Few-shot adaptation training.}
For each adaptation setting, we initialize the encoder and prediction
head from their corresponding source checkpoints and initialize new
input modules for the target sensors.

We use AdamW with a constant learning rate of $10^{-4}$ for the input
modules and, when the prediction head is trainable, $10^{-3}$ for the
head. The weight decay is $10^{-4}$, and the global batch size is 256,
balanced across the six adaptation sensor--object groups. We minimize
MSE on standardized six-dimensional wrenches using fixed source
normalization statistics.

At each data budget, let $N_{\mathrm{sf}}$ and $N_{\mathrm{tmp}}$
denote the numbers of single-frame samples and valid temporal windows,
respectively. We set the optimizer-step counts to
\begin{equation}
\begin{aligned}
T_{\mathrm{sf}}
&=\min\!\left(5000,
\left\lceil\frac{20N_{\mathrm{sf}}}{256}\right\rceil\right),\\
T_{\mathrm{tmp}}
&=\min\!\left(2500,
\left\lceil\frac{10N_{\mathrm{tmp}}}{256}\right\rceil\right).
\end{aligned}
\end{equation}
The factors 20 and 10 specify equivalent sample-draw budgets under
balanced minibatching. For single-frame training, the 5\%, 10\%, 25\%,
50\%, and 100\% data budgets correspond to 282, 563, 1,407, 2,813,
and 5,000 steps, respectively. Input-only adaptation uses the same
step-count rule for each mode at the 10\% data budget.

For each sensor and evaluation mode, we use a fixed set of 300
validation targets. Let $T$ denote the optimizer-step count for the
corresponding mode. We validate every $\lceil T/10\rceil$ steps and at
the final step. The checkpoint with the lowest sensor-balanced
normalized validation MSE is then evaluated on all eligible targets
from the test objects.

\paragraph{Detailed few-shot results.}
Table~\ref{tab:supp_fewshot_joint} reports normalized RMSE for
single-frame and temporal joint adaptation at all five data budgets.
Table~\ref{tab:supp_fewshot_input} reports input-only adaptation
results at the 10\% budget. Table~\ref{tab:supp_fewshot_summary}
summarizes performance across these conditions using the aggregation
schemes defined in Eq.~\ref{eq:supp_hierarchical_score} and the
accompanying text.

\begin{table*}[t]
    \centering
    \caption{
        Normalized RMSE ($\downarrow$) for joint adaptation
        of input modules and prediction heads on three unseen
        tactile sensors.
        Percentages indicate the fraction of the adaptation pool.
        The shared backbone remains frozen.
        Bold denotes the lowest error in each column
        within each evaluation setting.
    }
    \label{tab:supp_fewshot_joint}
    \small
    \setlength{\tabcolsep}{5pt}
    \renewcommand{\arraystretch}{1.12}
    \begin{tabular*}{\textwidth}{
        @{\extracolsep{\fill}}lccccc@{}
    }
        \toprule
        Method & 5\% & 10\% & 25\% & 50\% & 100\% \\
        \midrule
        \multicolumn{6}{c}{
            \textit{Single-frame joint adaptation}
        } \\
        \midrule
        DINOv2
        & \textbf{0.5453} & 0.5265 & 0.4943
        & \textbf{0.4486} & 0.4718 \\
        Sparsh
        & 0.5810 & 0.5483 & 0.5156 & 0.4932 & 0.4810 \\
        T3
        & 0.5805 & 0.5394 & \textbf{0.4841}
        & 0.4547 & \textbf{0.4224} \\
        AnyTouch2
        & 0.5457 & 0.5786 & 0.5026 & 0.4900 & 0.4773 \\
        HTT
        & 0.5458 & 0.5717 & 0.5432 & 0.5271 & 0.5180 \\
        Ours
        & 0.5539 & \textbf{0.5262} & 0.4992 & 0.4644 & 0.4668 \\
        \midrule
        \multicolumn{6}{c}{
            \textit{Temporal joint adaptation}
        } \\
        \midrule
        DINOv2
        & 0.4485 & 0.4014 & 0.3862 & 0.3862 & 0.3414 \\
        Sparsh
        & 0.4439 & 0.4123 & 0.3901 & 0.3767 & 0.3520 \\
        T3
        & 0.4339 & 0.4026 & 0.3818 & 0.3658 & 0.3328 \\
        AnyTouch2
        & 0.4254 & 0.4058 & 0.3957 & 0.3691 & 0.3486 \\
        HTT
        & \textbf{0.4120} & \textbf{0.3859}
        & 0.3767 & 0.3617 & 0.3381 \\
        Ours
        & 0.4187 & 0.3987 & \textbf{0.3736}
        & \textbf{0.3582} & \textbf{0.3294} \\
        \bottomrule
    \end{tabular*}
\end{table*}

\begin{table}[t]
    \centering
    \caption{
        Normalized RMSE ($\downarrow$) for input-only
        adaptation using 10\% of the adaptation pool.
        Both the shared backbone and prediction head
        remain frozen.
        Bold denotes the lowest error in each column.
    }
    \label{tab:supp_fewshot_input}
    \small
    \setlength{\tabcolsep}{5pt}
    \renewcommand{\arraystretch}{1.12}
    \begin{tabular*}{\linewidth}{
        @{\extracolsep{\fill}}lcc@{}
    }
        \toprule
        Method & Single-frame & Temporal \\
        \midrule
        DINOv2    & 0.5103 & 0.4238 \\
        Sparsh    & 0.5832 & 0.4163 \\
        T3        & 0.5311 & 0.4079 \\
        AnyTouch2 & 0.5905 & 0.4423 \\
        HTT       & 0.6414 & \textbf{0.4030} \\
        Ours      & \textbf{0.5046} & 0.4203 \\
        \bottomrule
    \end{tabular*}
\end{table}

\paragraph{Aggregate Results.}
Table~\ref{tab:supp_fewshot_summary} reports the hierarchical
aggregate scores defined in
Eq.~\ref{eq:supp_hierarchical_score}, together with
an equal-weight average over all 12 conditions.
Our method achieves the lowest aggregate error under
both schemes.
Performance varies across individual settings:
our method achieves the lowest temporal joint-adaptation
error at the 25\%, 50\%, and 100\% budgets, while HTT
performs best at 5\% and 10\%.
Our method also achieves the lowest single-frame
input-only adaptation error.

\begin{table}[t]
    \centering
    \caption{
        Aggregate few-shot normalized RMSE ($\downarrow$).
        Hierarchical scores weight the three adaptation
        settings equally; equal-weight scores average
        all 12 conditions equally.
        Values are computed before rounding to four
        decimal places.
    }
    \label{tab:supp_fewshot_summary}
    \small
    \setlength{\tabcolsep}{4pt}
    \renewcommand{\arraystretch}{1.12}
    \begin{tabular*}{\linewidth}{
        @{\extracolsep{\fill}}lcc@{}
    }
        \toprule
        Method
        & \shortstack{Hierarchical\\score}
        & \shortstack{Equal-weight\\mean} \\
        \midrule
        DINOv2    & 0.4524 & 0.4487 \\
        Sparsh    & 0.4729 & 0.4661 \\
        T3        & 0.4497 & 0.4448 \\
        AnyTouch2 & 0.4747 & 0.4643 \\
        HTT       & 0.4794 & 0.4687 \\
        Ours      & \textbf{0.4468} & \textbf{0.4428} \\
        \bottomrule
    \end{tabular*}
\end{table}

\subsection{Real-World Task Definitions}
\label{sec:supp_real_tasks}

We evaluate four real-world manipulation tasks involving
stable grasping, precise alignment, or sustained surface contact.
The task objectives and success criteria are defined as follows.

\paragraph{Lift Bottle.}
The robot grasps a bottle and lifts it from the supporting
surface, requiring a stable grasp to prevent slippage.
\textit{Success:} The bottle is lifted clear of the surface
without being dropped.

\paragraph{Remove Power Bank.}
The robot grasps a power bank and removes it from its initial
position, requiring a stable grasp throughout the removal.
\textit{Success:} The power bank is fully removed from its
initial position without being dropped.

\paragraph{Key Insertion.}
The robot grasps a key, aligns it with the keyhole, and inserts
it, requiring precise alignment and controlled contact.
\textit{Success:} The key is successfully inserted into the keyhole.

\paragraph{Board Cleaning.}
The robot moves a cleaning tool across a board to remove
designated marks, requiring sustained tool-board contact
throughout the cleaning motion.
\textit{Success:} All designated marks are removed.

\section{Ablation and Design Comparison Settings}
\label{sec:supp_ablation}

This section describes the ablation settings used to assess the
components of our framework. Retrieval follows the protocol in
Appendix~\ref{sec:supp_retrieval}.

\paragraph{Sensor-specific MLPs.}
To assess the contribution of sensor-specific adaptation for
image-based inputs, we remove the sensor-specific residual MLPs
applied to image patch tokens. We retain the shared patch projection
and positional embeddings, vector-valued MLP tokenizers,
pressure-matrix convolutional frontend, and sensor identity tokens.
All other model components and training settings remain unchanged.
This variant is labeled ``w/o Sensor-specific MLPs'' in the main
ablation table.

\paragraph{Direct wrench supervision.}
To evaluate the benefit of wrench-grounded representation learning,
we compare the full model with a baseline that trains the tactile
encoder to predict wrenches directly from tactile features. 

The baseline retains the same sensor input modules, shared Transformer,
and physical-query pooling as the full model. Its regression head maps
the pooled feature $\mathbf{h}\in\mathbb{R}^{768}$ to the standardized
wrench through LayerNorm, Linear $768\rightarrow256$, GELU, and
Linear $256\rightarrow6$. The tactile encoder and regression head are
jointly trained using mean squared error on the standardized wrench.
For frozen-probe and retrieval evaluation, we use $\mathbf{h}$
directly as the tactile representation.

Unlike the full model, this baseline omits the
$\mathbf{h}$-to-$\mathbf{z}$ projection, wrench encoder, EMA targets,
and representation alignment loss. Training data, optimization budgets,
and downstream evaluation protocols are otherwise identical to those
of the full model. This baseline is labeled ``Direct wrench
supervision'' in the main ablation table.

\paragraph{Alignment loss.}
To assess the contribution of aligning tactile representations with
wrench representations, we set the weight of the alignment loss to
zero while retaining both wrench reconstruction losses and their
weights. The EMA wrench encoder remains in the model and continues
to be updated, but its output no longer contributes to the training
objective. This variant is labeled ``w/o Alignment'' in the main
ablation table.

\paragraph{Sensor identity token.}
To assess the contribution of explicitly providing sensor identity,
we remove the sensor identity token from the tactile input sequence.
We retain the sensor-specific adapters, positional embeddings,
pretrained CLS token, and physical query. This variant is
labeled ``w/o Sensor Identity Token'' in the main ablation table.

\begin{thebibliography}{38}
\providecommand{\natexlab}[1]{#1}
\providecommand{\url}[1]{\texttt{#1}}
\expandafter\ifx\csname urlstyle\endcsname\relax
  \providecommand{\doi}[1]{doi: #1}\else
  \providecommand{\doi}{doi: \begingroup \urlstyle{rm}\Url}\fi

\bibitem[Bhirangi et~al.(2025)Bhirangi, Pattabiraman, Erciyes, Cao,
  Hellebrekers, and Pinto]{bhirangi2025anyskin}
Raunaq Bhirangi, Venkatesh Pattabiraman, Enes Erciyes, Yifeng Cao, Tess
  Hellebrekers, and Lerrel Pinto.
\newblock Anyskin: Plug-and-play skin sensing for robotic touch.
\newblock In \emph{2025 IEEE International Conference on Robotics and
  Automation (ICRA)}, pages 16563--16570. IEEE, 2025.

\bibitem[Bi et~al.(2026)Bi, Wang, Reddy, Lin, Khajikhanov, Gao, and
  Soh]{bi2026heterogeneous}
Jianxin Bi, Qiang Wang, Jayaram Reddy, Kelvin Lin, Soibkhon Khajikhanov, Ruihan
  Gao, and Harold Soh.
\newblock Heterogeneous tactile transformer.
\newblock \emph{arXiv preprint arXiv:2606.29948}, 2026.

\bibitem[Billard and Kragic(2019)]{billard2019trends}
Aude Billard and Danica Kragic.
\newblock Trends and challenges in robot manipulation.
\newblock \emph{Science}, 364\penalty0 (6446):\penalty0 eaat8414, 2019.

\bibitem[Chen et~al.(2026{\natexlab{a}})Chen, Wan, Chen, Guo, Xu, Qi, Zhang,
  Wu, Xu, Li, et~al.]{chen2026univtac}
Baijun Chen, Weijie Wan, Tianxing Chen, Xianda Guo, Congsheng Xu, Yuanyang Qi,
  Haojie Zhang, Longyan Wu, Tianling Xu, Zixuan Li, et~al.
\newblock Univtac: A unified simulation platform for visuo-tactile manipulation
  data generation, learning, and benchmarking.
\newblock \emph{arXiv preprint arXiv:2602.10093}, 2026{\natexlab{a}}.

\bibitem[Chen et~al.(2026{\natexlab{b}})Chen, Ni, Luo, Wu, Zhang,
  Spyrakos-Papastavridis, Jamone, Lepora, Deng, and Luo]{chen2026uniforce}
Zhuo Chen, Fei Ni, Kaiyao Luo, Zhiyuan Wu, Xuyang Zhang, Emmanouil
  Spyrakos-Papastavridis, Lorenzo Jamone, Nathan~F Lepora, Jiankang Deng, and
  Shan Luo.
\newblock Uniforce: A unified latent force model for robot manipulation with
  diverse tactile sensors.
\newblock \emph{arXiv preprint arXiv:2602.01153}, 2026{\natexlab{b}}.

\bibitem[Cheng et~al.(2026)Cheng, Xu, Li, Chen, Gao, Wang, Peng, and
  Han]{cheng2026tau}
Ning Cheng, Jinan Xu, Wanlin Li, Yangzhi Chen, Jing Gao, Yiqun Wang, Kelan
  Peng, and Wenjuan Han.
\newblock {$\tau$}: Learning touch-augmented vision-language-action models from
  future visual supervision.
\newblock \emph{arXiv preprint arXiv:2607.24485}, 2026.

\bibitem[Feng et~al.(2026)Feng, Zhou, Mei, Zhou, Wang, Cui, Fang, Yao, and
  Hu]{feng2026anytouch}
Ruoxuan Feng, Yuxuan Zhou, Siyu Mei, Dongzhan Zhou, Pengwei Wang, Shaowei Cui,
  Bin Fang, Guocai Yao, and Di Hu.
\newblock Anytouch 2: General optical tactile representation learning for
  dynamic tactile perception.
\newblock \emph{arXiv preprint arXiv:2602.09617}, 2026.

\bibitem[He et~al.(2026)He, F{\"a}rber, and Calandra]{he2026rct}
Jingbo He, Michael F{\"a}rber, and Roberto Calandra.
\newblock Rct: A robot-collected touch-vision-language dataset for tactile
  generalization.
\newblock \emph{arXiv preprint arXiv:2606.31694}, 2026.

\bibitem[Helmut et~al.(2026)Helmut, Funk, Schneider, de~Farias, and
  Peters]{helmut2026tactile}
Erik Helmut, Niklas Funk, Tim Schneider, Cristiana de Farias, and Jan Peters.
\newblock Tactile-conditioned diffusion policy for force-aware robotic
  manipulation.
\newblock In \emph{2026 IEEE International Conference on Robotics and
  Automation (ICRA)}, pages 1815--1822. IEEE, 2026.

\bibitem[Higuera et~al.(2024)Higuera, Sharma, Bodduluri, Fan, Lancaster,
  Kalakrishnan, Kaess, Boots, Lambeta, Wu, et~al.]{higuera2024sparsh}
Carolina Higuera, Akash Sharma, Chaithanya~Krishna Bodduluri, Taosha Fan,
  Patrick Lancaster, Mrinal Kalakrishnan, Michael Kaess, Byron Boots, Mike
  Lambeta, Tingfan Wu, et~al.
\newblock Sparsh: Self-supervised touch representations for vision-based
  tactile sensing.
\newblock \emph{arXiv preprint arXiv:2410.24090}, 2024.

\bibitem[Hou et~al.(2025)Hou, Sun, Wang, Huang, Zhang, Liu, Li, and
  Jiao]{hou2025supermag}
Peiyao Hou, Danning Sun, Meng Wang, Yuzhe Huang, Zeyu Zhang, Hangxin Liu,
  Wanlin Li, and Ziyuan Jiao.
\newblock Supermag: Vision-based tactile data guided high-resolution tactile
  shape reconstruction for magnetic tactile sensors.
\newblock In \emph{2025 IEEE/RSJ International Conference on Intelligent Robots
  and Systems (IROS)}, pages 8561--8567. IEEE, 2025.

\bibitem[Huang and Li(2026)]{huang2026flexitac}
Binghao Huang and Yunzhu Li.
\newblock Flexitac: A low-cost, open-source, scalable tactile sensing solution
  for robotic systems.
\newblock \emph{arXiv preprint arXiv:2604.28156}, 2026.

\bibitem[Huang et~al.(2025)Huang, Xu, and Xiao]{huang2025twintac}
Xiyan Huang, Zhe Xu, and Chenxi Xiao.
\newblock Twintac: A wide-range, highly sensitive tactile sensor with
  real-to-sim digital twin sensor model.
\newblock In \emph{2025 IEEE/RSJ International Conference on Intelligent Robots
  and Systems (IROS)}, pages 12286--12292. IEEE, 2025.

\bibitem[Huang et~al.(2026{\natexlab{a}})Huang, Li, and Jiao]{huang2026tactile}
Yuzhe Huang, Wanlin Li, and Ziyuan Jiao.
\newblock Tactile-guided exploration and positioning for high-precision robotic
  peg-in-hole tasks.
\newblock \emph{IEEE/ASME Transactions on Mechatronics}, 2026{\natexlab{a}}.

\bibitem[Huang et~al.(2026{\natexlab{b}})Huang, Lin, Li, Li, Li, Jiang, Xiao,
  and Jiao]{huang2026taf}
Yuzhe Huang, Pei Lin, Wanlin Li, Daohan Li, Jiajun Li, Jiaming Jiang, Chenxi
  Xiao, and Ziyuan Jiao.
\newblock Taf-vla: Tactile-force alignment in vision-language-action models for
  force-aware manipulation.
\newblock \emph{arXiv preprint arXiv:2601.20321}, 2026{\natexlab{b}}.

\bibitem[Huang et~al.(2026{\natexlab{c}})Huang, Wu, Jiang, Lin, Aierken, Wang,
  Cheng, Li, Xiao, Jiao, et~al.]{huang2026ht}
Yuzhe Huang, Jiaping Wu, Jiaming Jiang, Hezhe Lin, Aikebaier Aierken, Yunlong
  Wang, Kun Cheng, Wanlin Li, Chenxi Xiao, Ziyuan Jiao, et~al.
\newblock Ht-bench: Benchmarking and learning dexterous full-hand tactile
  representations with egocentric vision.
\newblock \emph{arXiv preprint arXiv:2606.19161}, 2026{\natexlab{c}}.

\bibitem[Jiang et~al.(2026)Jiang, Huang, Liang, Lin, Luo, Dong, Wu, Xiao, Li,
  and Jiao]{jiang2026caat}
Jiaming Jiang, Yuzhe Huang, Hao Liang, Pei Lin, Shengcheng Luo, Fanrong Dong,
  Jiaping Wu, Chenxi Xiao, Wanlin Li, and Ziyuan Jiao.
\newblock Caat: Contact-aware attention scaling and tactile masking for
  data-efficient contact-rich manipulation.
\newblock \emph{arXiv preprint arXiv:2608.01102}, 2026.

\bibitem[Lambeta et~al.(2020)Lambeta, Chou, Tian, Yang, Maloon, Most, Stroud,
  Santos, Byagowi, Kammerer, et~al.]{lambeta2020digit}
Mike Lambeta, Po-Wei Chou, Stephen Tian, Brian Yang, Benjamin Maloon,
  Victoria~Rose Most, Dave Stroud, Raymond Santos, Ahmad Byagowi, Gregg
  Kammerer, et~al.
\newblock Digit: A novel design for a low-cost compact high-resolution tactile
  sensor with application to in-hand manipulation.
\newblock \emph{IEEE Robotics and Automation Letters}, 5\penalty0 (3):\penalty0
  3838--3845, 2020.

\bibitem[Li et~al.(2020)Li, Kroemer, Su, Veiga, Kaboli, and
  Ritter]{li2020review}
Qiang Li, Oliver Kroemer, Zhe Su, Filipe~Fernandes Veiga, Mohsen Kaboli, and
  Helge~Joachim Ritter.
\newblock A review of tactile information: Perception and action through touch.
\newblock \emph{IEEE Transactions on Robotics}, 36\penalty0 (6):\penalty0
  1619--1634, 2020.

\bibitem[Lin et~al.(2023)Lin, Zhang, Xu, Wu, and Xu]{lin20239dtact}
Changyi Lin, Han Zhang, Jikai Xu, Lei Wu, and Huazhe Xu.
\newblock 9dtact: A compact vision-based tactile sensor for accurate 3d shape
  reconstruction and generalizable 6d force estimation.
\newblock \emph{IEEE Robotics and Automation Letters}, 9\penalty0 (2):\penalty0
  923--930, 2023.

\bibitem[Lin et~al.(2025)Lin, Huang, Li, Ma, Xiao, and Jiao]{lin2025pp}
Pei Lin, Yuzhe Huang, Wanlin Li, Jianpeng Ma, Chenxi Xiao, and Ziyuan Jiao.
\newblock Pp-tac: Paper picking using tactile feedback in dexterous robotic
  hands.
\newblock \emph{arXiv preprint arXiv:2504.16649}, 2025.

\bibitem[Lin et~al.(2026)Lin, Huang, Li, Xiao, and Jiao]{lin2026dexmove}
Pei Lin, Yuzhe Huang, Wanlin Li, Chenxi Xiao, and Ziyuan Jiao.
\newblock Dexmove: Learning tactile-guided non-prehensile manipulation with
  dexterous hands.
\newblock In \emph{The Fourteenth International Conference on Learning
  Representations}, 2026.

\bibitem[Ma et~al.(2026{\natexlab{a}})Ma, Cao, Zhang, Zhang, Yang, Hao, Fang,
  Cai, Cui, and Wang]{ma2026cltp}
Wenxuan Ma, Xiaoge Cao, Yixiang Zhang, Chaofan Zhang, Shaobo Yang, Peng Hao,
  Bin Fang, Yinghao Cai, Shaowei Cui, and Shuo Wang.
\newblock Cltp: Contrastive language-tactile pre-training for 3d contact
  geometry understanding.
\newblock \emph{Biomimetic Intelligence and Robotics}, page 100324,
  2026{\natexlab{a}}.

\bibitem[Ma et~al.(2026{\natexlab{b}})Ma, Zhang, Cai, Yao, Cui, and
  Wang]{ma2026fg}
Wenxuan Ma, Chaofan Zhang, Yinghao Cai, Guocai Yao, Shaowei Cui, and Shuo Wang.
\newblock Fg-cltp: Fine-grained contrastive language tactile pretraining for
  robotic manipulation.
\newblock \emph{IEEE Robotics and Automation Letters}, 2026{\natexlab{b}}.

\bibitem[Mao et~al.(2024)Mao, Liao, Yuan, and Zhu]{mao2024multimodal}
Qian Mao, Zijian Liao, Jinfeng Yuan, and Rong Zhu.
\newblock Multimodal tactile sensing fused with vision for dexterous robotic
  housekeeping.
\newblock \emph{Nature Communications}, 15\penalty0 (1):\penalty0 6871, 2024.

\bibitem[Niu et~al.(2026)Niu, Liu, Wang, Shao, Yin, Pai, Sharma, Saravalle,
  Zheng, Wang, et~al.]{niu2026t}
Dantong Niu, Zhuoyang Liu, Zekai Wang, Boning Shao, Zhao-Heng Yin, Anirudh Pai,
  Yuvan Sharma, Stefano Saravalle, Ruijie Zheng, Jing Wang, et~al.
\newblock T-rex: Tactile-reactive dexterous manipulation.
\newblock \emph{arXiv preprint arXiv:2606.17055}, 2026.

\bibitem[Oquab et~al.(2023)Oquab, Darcet, Moutakanni, Vo, Szafraniec, Khalidov,
  Fernandez, Haziza, Massa, El-Nouby, et~al.]{oquab2023dinov2}
Maxime Oquab, Timoth{\'e}e Darcet, Th{\'e}o Moutakanni, Huy Vo, Marc
  Szafraniec, Vasil Khalidov, Pierre Fernandez, Daniel Haziza, Francisco Massa,
  Alaaeldin El-Nouby, et~al.
\newblock Dinov2: Learning robust visual features without supervision.
\newblock \emph{arXiv preprint arXiv:2304.07193}, 2023.

\bibitem[Park et~al.(2026)Park, Bhadang, Sferrazza, Yi, and
  Wang]{park2026tactx}
Junsung Park, Sachin Bhadang, Carmelo Sferrazza, Sha Yi, and Xiaolong Wang.
\newblock Tactx: Learning shared tactile representations across diverse
  sensors.
\newblock \emph{arXiv preprint arXiv:2606.31236}, 2026.

\bibitem[Qin et~al.(2026)Qin, Zhao, Leng, Hu, and Xiao]{qin2026nlipscalib}
Xuhao Qin, Feiyu Zhao, Yatao Leng, Runze Hu, and Chenxi Xiao.
\newblock Nlipscalib: An efficient calibration framework for high-fidelity 3d
  reconstruction of curved visuotactile sensors.
\newblock \emph{arXiv preprint arXiv:2603.09319}, 2026.

\bibitem[Team et~al.(2026)Team, Team, et~al.]{team2026n_0}
NeoteAI Team, Fudan~TEAI Team, et~al.
\newblock $ n\_0 $-foundation: Towards the age of tactile intelligence.
\newblock \emph{arXiv preprint arXiv:2608.29601}, 2026.

\bibitem[Van Den~Oord et~al.(2017)Van Den~Oord, Vinyals, et~al.]{van2017neural}
Aaron Van Den~Oord, Oriol Vinyals, et~al.
\newblock Neural discrete representation learning.
\newblock \emph{Advances in neural information processing systems}, 30, 2017.

\bibitem[Wu et~al.(2025)Wu, Chen, Zhu, Qin, and Xiao]{wu2025humanft}
Yifan Wu, Yuzhou Chen, Zhengying Zhu, Xuhao Qin, and Chenxi Xiao.
\newblock Humanft: A human-like fingertip multimodal visuo-tactile sensor.
\newblock In \emph{2025 IEEE International Conference on Robotics and
  Automation (ICRA)}, pages 244--250. IEEE, 2025.

\bibitem[Yuan et~al.(2017)Yuan, Dong, and Adelson]{yuan2017gelsight}
Wenzhen Yuan, Siyuan Dong, and Edward~H Adelson.
\newblock Gelsight: High-resolution robot tactile sensors for estimating
  geometry and force.
\newblock \emph{Sensors}, 17\penalty0 (12):\penalty0 2762, 2017.

\bibitem[Zhang et~al.(2023)Zhang, Cui, Wang, Hu, Cai, Wang, and
  Wang]{zhang2023gelstereo}
Chaofan Zhang, Shaowei Cui, Shuo Wang, Jingyi Hu, Yinghao Cai, Rui Wang, and Yu
  Wang.
\newblock Gelstereo 2.0: An improved gelstereo sensor with multimedium
  refractive stereo calibration.
\newblock \emph{IEEE Transactions on Industrial Electronics}, 71\penalty0
  (7):\penalty0 7452--7462, 2023.

\bibitem[Zhao et~al.(2024{\natexlab{a}})Zhao, Ma, Wang, and
  Adelson]{zhao2024transferable}
Jialiang Zhao, Yuxiang Ma, Lirui Wang, and Edward~H Adelson.
\newblock Transferable tactile transformers for representation learning across
  diverse sensors and tasks.
\newblock \emph{arXiv preprint arXiv:2406.13640}, 2024{\natexlab{a}}.

\bibitem[Zhao et~al.(2023)Zhao, Kumar, Levine, and Finn]{zhao2023learning}
Tony~Z Zhao, Vikash Kumar, Sergey Levine, and Chelsea Finn.
\newblock Learning fine-grained bimanual manipulation with low-cost hardware.
\newblock \emph{arXiv preprint arXiv:2304.13705}, 2023.

\bibitem[Zhao et~al.(2024{\natexlab{b}})Zhao, Li, Li, Qi, Ruan, Zhu, and
  Althoefer]{zhao2024tac}
Zihang Zhao, Yuyang Li, Wanlin Li, Zhenghao Qi, Lecheng Ruan, Yixin Zhu, and
  Kaspar Althoefer.
\newblock Tac-man: Tactile-informed prior-free manipulation of articulated
  objects.
\newblock \emph{IEEE Transactions on Robotics}, 41:\penalty0 538--557,
  2024{\natexlab{b}}.

\bibitem[Zhao et~al.(2025)Zhao, Li, Li, Liu, Li, Wang, Du, Liu, Zhu, Wang,
  et~al.]{zhao2025embedding}
Zihang Zhao, Wanlin Li, Yuyang Li, Tengyu Liu, Boren Li, Meng Wang, Kai Du,
  Hangxin Liu, Yixin Zhu, Qining Wang, et~al.
\newblock Embedding high-resolution touch across robotic hands enables adaptive
  human-like grasping.
\newblock \emph{Nature Machine Intelligence}, 7\penalty0 (6):\penalty0
  889--900, 2025.

\end{thebibliography}
\end{document}